%% file: 00-main.tex
\pdfoutput=1
\documentclass[11pt]{article}

\usepackage{acl}

\usepackage{times}
\usepackage{latexsym}
\usepackage[T1]{fontenc}
\usepackage[utf8]{inputenc}
\usepackage{microtype}
\usepackage{inconsolata}

\usepackage{graphicx}
\usepackage{times}
\usepackage{fancyhdr,graphicx,amsmath,amssymb}
\usepackage[ruled,vlined]{algorithm2e}
\usepackage{multirow}
\usepackage{amsfonts}
\usepackage{colortbl}
\usepackage{hyperref}
\usepackage{mwe}
\usepackage{makecell}
\usepackage{tikz}
\usepackage{listings}
\usepackage{xspace}
\usepackage{hhline}
\usepackage{tabularray}
\usepackage[export]{adjustbox}
\usepackage{float}
\usepackage{dblfloatfix}

\newcommand{\GAME}{\texttt{TRAVELING ADVENTURERS}\xspace}

\newcommand{\action}[1]{\texttt{#1}}

\newcommand{\acts}{\action{suggest}\xspace}
\newcommand{\actag}{\action{agree}\xspace}
\newcommand{\actai}{\action{agree-inner}\xspace}
\newcommand{\actao}{\action{agree-outer}\xspace}
\newcommand{\actr}{\action{reject}\xspace}
\newcommand{\acti}{\action{inform}\xspace}
\newcommand{\actas}{\action{ask}\xspace}
\newcommand{\actv}{\action{visit}\xspace}
\newcommand{\acte}{\action{end}\xspace}

\newcommand{\USER}{\texttt{USER}\xspace}
\newcommand{\BOT}{\texttt{BOT}\xspace}

\title{Collaborative Problem-Solving in an Optimization Game}

\author{Isidora Jeknić \\
  Saarland University \\
  Saarbrücken, Germany\\
  \footnotesize{{\tt jeknic@lst.uni-saarland.de}
  }
  \\\And
  Alex Duchnowski \\
  Saarland University \\
  Saarbrücken, Germany\\
  \footnotesize{{\tt aduchnowski@coli.uni-saarland.de}}
  \\\And
  Alexander Koller \\
  Saarland University \\
  Saarbrücken, Germany\\
  \footnotesize{{\tt koller@coli.uni-saarland.de}}
  }

\begin{document}
\maketitle
\begin{abstract}
    Dialogue agents that support human users in solving complex tasks have received much attention recently.
    Many such tasks are NP-hard optimization problems that require careful collaborative exploration of the solution space.
    We introduce a novel dialogue game in which the agents collaboratively solve a two-player Traveling Salesman problem,
    along with an agent that combines LLM prompting with symbolic mechanisms for state tracking and grounding.
    Our best agent solves 45\% of games optimally in self-play. 
    It also demonstrates an ability to collaborate successfully with human users and generalize to unfamiliar graphs.

\end{abstract}

\input{01-introduction}
\input{02-background}
\input{03-game}
\input{04-versions}
\input{05-results}
\input{06-conclusion}

\section*{Limitations}
The results reflect performance on a 6-node graph only, which is also the size of the prompt example game. During development, we also tested 4- and 5-node graphs, but the 6-node version proved to be much more difficult and thus more interesting. In the future, it would be informative to compare performance across larger graph sizes. In order to assure robustness, it would be interesting to conduct further experiments of the model on larger graphs.
Similarly, the presented game only represents one optimization problem, namely the TSP. It would be informative to see if the agent structure from Section \ref{sec:vers} is applicable to other problems.

Additionally, we find the agents prefer an incremental and greedy node-by-node problem solving approach, which may not result in a true optimal (but instead only a greedy optimal) solution. 
It would be insightful to run the experiment on greedy-only boards and observe the difference in results.

Regarding the human-agent experiments, there were games in which the human player explained an alternative path from a previously agreed-upon node. A limitation of the agent design is that it has no backtracking abilities, which means that the internal symbolic state-tracking structures cannot adjust previous agreement. However, they can still revise a complete path.

Another limitation is the omission of human-human data, which makes it difficult to adequately assess and compare the performance of the human-agent setup. Additionally, the difficulty of the task is typically determined based on how long it takes humans to complete the task and how successful they are. With no frame of reference regarding human performance, this becomes harder to do.

Lastly, another pertinent next step would be to evaluate the performance of open source models such as Llama on the task, which would allow for reliable reproducibility.

\section*{Ethics Statement}
We do not see any particular ethical challenges with
the research reported here.

\section*{Acknowledgments}
We thank Florian Kandra for his help with setting up the Slurk interface. This research was partially funded by the Deutsche Forschungsgemeinschaft (DFG, German Research Foundation) – Project-ID KO 2916/3-1.

\bibliography{00-main}

\newpage
\appendix
\input{appendix}

\end{document}

%% file: 01-introduction.tex
% !TeX root = main.tex

\section{Introduction}
\label{sec:intro}

Humans frequently face the challenge of solving hard combinatorial problems in their daily lives,
from planning to scheduling to resource allocation, and they often struggle to solve these problems well
\citep{hidalgo13}.
LLMs are surprisingly good at solving these problems \cite{fan-etal-2024-nphardeval}, but
only when the human user has full knowledge of the problem and can spell it out in detail.
One of the great promises of LLMs is that they can help humans solve such problems \emph{collaboratively} \cite{khan2024capability}, through a dialogue in which the human and the system
take turns working out the problem and proposing increasingly complete and correct solutions.

AI agents for collaborative problem-solving must have a number of fundamental skills to be effective.
In addition to the ability to solve NP-hard problems themselves, they also need to perform conversational grounding
(tracking the details of the problem and which part of the solution have we agreed on), remember what their partner
wants and knows, and negotiate and revise partial solutions. All of these are established problems in the dialogue
literature, and it is not obvious that LLMs will solve them easily.

In this paper, we make two contributions to the development of collaborative agents for solving
complex problems.\footnote{Our code and data are available at: \newline \hyperlink{https://github.com/coli-saar/collaborative-problem-solving}{https://github.com/coli-saar/collaborative-problem-solving}.} First, we introduce \GAME , a game in which two agents collaboratively solve a Traveling Salesman problem.
This game serves as a testbed for collaborative problem-solving, as each agent only has partial information
about the problem initially and must communicate and collaborate in order to negotiate an optimal solution.
Furthermore, because it builds on an NP-hard optimization problem, \GAME\ is difficult enough that
humans could not immediately find optimal solutions even with access to their partner's knowledge,
necessitating the incremental construction of a solution that we expect in a collaborative problem-solving dialogue.

Second, we present an artificial agent that performs well on playing \GAME . We find that a purely LLM-based baseline
struggles with tracking its partner's knowledge and with conversational grounding; it is outperformed decisively
by a neurosymbolic agent which combines the LLM with symbolic components for these skills and an exact symbolic optimizer.
The neurosymbolic agent solves 98\% of its games correctly and 45\% optimally in self-play with itself, and still achieves a high optimality
score in 32\% of games with human partners. We conclude with a discussion of conversational skills
that would be required to push the optimality rate even higher.

\begin{figure*}[h!]
    \centering
    \includegraphics[width=1\linewidth]{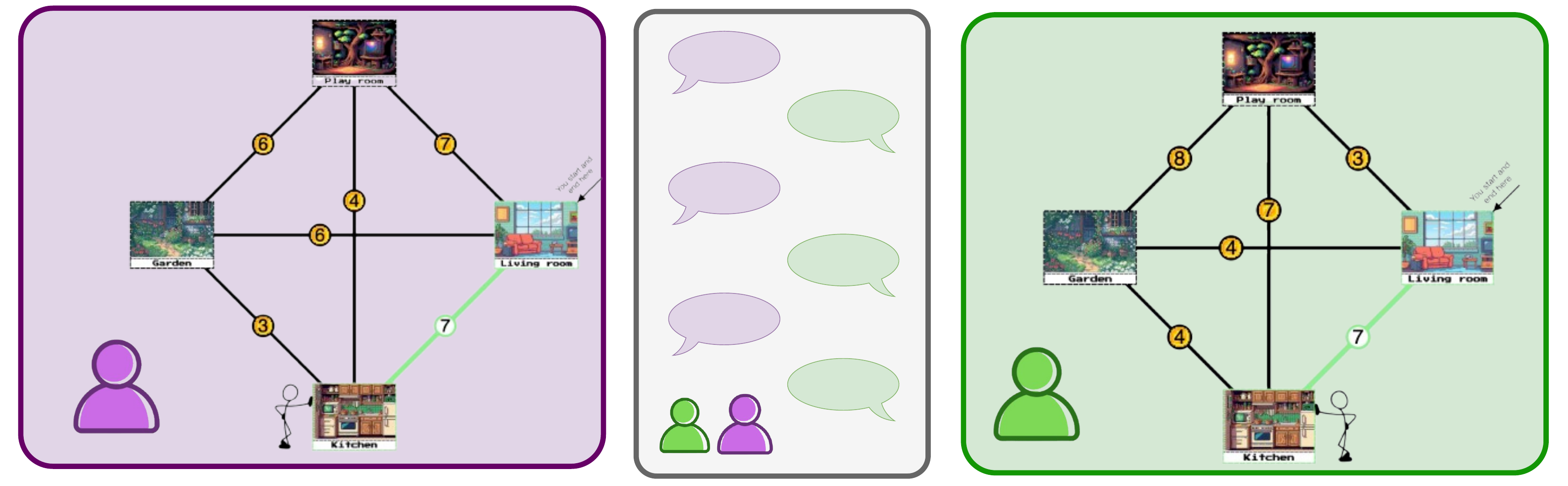}
    \caption{The perspectives of the two players - the left and right figures depict the distinct graphs of each player, whereas the middle illustrates the shared information, i.e., the chat.}
    \label{fig:view}
\end{figure*}

%% file: 02-background.tex
% !TeX root = main.tex
\section{Background}
\label{sec:bg}

In order to facilitate collaboration, interlocutors must engage in dialogue so as to ensure a shared set of intentions \cite{dafoe2021cooperative}.
Collaboration in dialogue is realized through negotiation, i.e., effort to establish common ground and a joint purpose \cite{clark_1996_using}. It is often identified through a set of dialogue acts, defined as ``meaning[s] of [utterances] at the level of illocutionary force'' \citep[2]{stolcke2000dialogue}, such as \textbf{presenting} a solution, as well as \textbf{accepting} or \textbf{rejecting} a proposal \cite{georgila2011learning}. Therefore, a collaborative agent must successfully process and produce such acts.

Task-oriented collaborative dialogue is primarily studied through \textit{collaborative games} which foster negotiation by incentivizing goal-driven conversation \cite{schlangen2019groundedagreementgamesemphasizing}. This often includes simple problem-solving tasks
% , such as describing a fixed image 
\cite{zarrie_2016_pentoref, lewis2017deal, kim-etal-2019-codraw}. Most of these environments---even those which are more complex, such as TRAINS \cite{doi:10.1080/09528139508953799}---are typically asymmetric, assigning players fixed roles. This limits the scope of collaboration that can be studied and prevents organic role-taking, despite human players preferring a more balanced role distribution, which often improves performance on symmetric tasks \cite{jeknic-etal-2024-dialogue}. 

LLMs have shown promising results and further potential on very complex tasks (for overview, see \citet{cheng2024exploringlargelanguagemodel,mialon2023augmented, hartmann2024surveycomplextasksgoaldirected}). However, previous work has shown that even the most advanced models (at the time, GPT-4) severely under-performs humans on strategic thinking tasks \cite{zhou2024sotopiainteractiveevaluationsocial}. Similarly, when dealing with complex tasks, human input can be beneficial for agent performance and make up for certain model-specific failures \cite{lin-etal-2024-decision}. 
Additionally, most previous work investigating LLMs and NP-hard problems is focused on single-agent optimization \cite{ramamonjison2023nl4opt, tang2025grapharenaevaluatingexploringlarge}.

%% file: 03-game.tex
% !TeX root = main.tex

\section{The Game}
\label{sec:game}

\begin{figure*}
    \centering
    \includegraphics[width=1\linewidth]{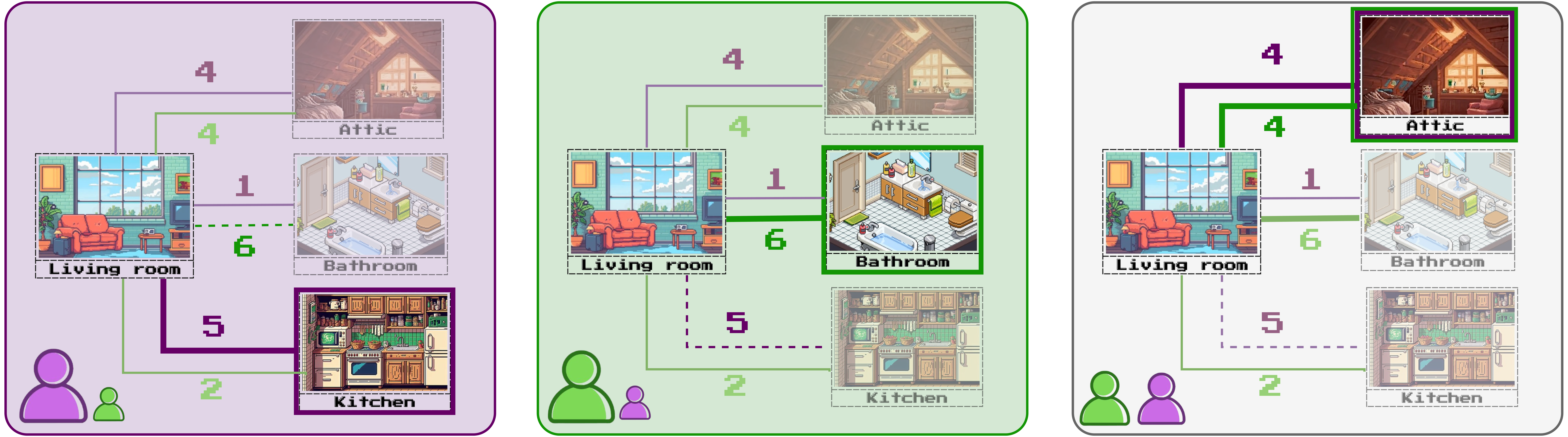}
    \caption{Different perspectives of the same move: the \textbf{left} and \textbf{middle} panels show two players' individual greedy optimal move; the \textbf{right} panel shows the joint greedy optimal move, which does not match either player's individual optimum.}
    \label{fig:decisions}
\end{figure*}

We present \GAME, a symmetric two-player game environment for collecting task-oriented collaborative dialogue. Players A and B observe a fully connected undirected graph $G$: the set of nodes ($V$) and edges ($E$) are identical for both players, whereas the weights ($w$) are unique ($w_A, w_B$) and only known to the respective player. The goal is to find a single path that visits all nodes exactly once and returns to the starting node while maximizing the sum of players' weights along the way. The pairs communicate exclusively through written text and have access to the entire chat history throughout the game.

\paragraph{Story} The players are told that an evil wizard imprisoned them in two parallel dimensions of the same house: the \textit{light} and \textit{ghost} dimensions. They observe the same layout, where each room is connected to all the others via hallways. In order to break the curse, the players are instructed to visit each room together and return to the start. Moreover, each player collects their respective type of coin (the \textit{light} and \textit{ghost} coins) in the hallways between rooms. In the end, the sum of coins collected by each player are given to them as joint loot.

\subsection{Features}

The game mechanics are based on a classical NP-hard optimization problem, the Traveling Salesman Problem (TSP), which we adjust for two players and ``dress up'' \cite{duchnowski2025ehopdataseteverydaynphard} into the familiar game setting of collecting coins. In relation to the TSP, the rooms represent cities and the coins distances (Figure \ref{fig:view}), with the coin maximization analogous to distance minimization. The curse-breaking ritual follows the rules of the TSP, and the coin reward in the end measures the overall optimality of the selected path. Because of this setup, with a high enough number of rooms it is not immediately clear to humans what an optimal solution is, which necessitates problem-solving.

The initial setup is symmetric, since both players are each given the same amount and type of information: their own world state is fully known, and the other player's world state is partially known.
The agents are not assigned predetermined roles and are free to individually choose an approach.

Because the same decision might not be optimal for both players individually, no single player is able to find an optimal solution alone (Figure \ref{fig:decisions}). This produces the need for negotiation.

%% file: 04-versions.tex
% !TeX root = main.tex

\section{Game-Playing Agents}
\label{sec:vers}

\begin{figure*}[t]
    \centering
    \includegraphics[width=1\linewidth]{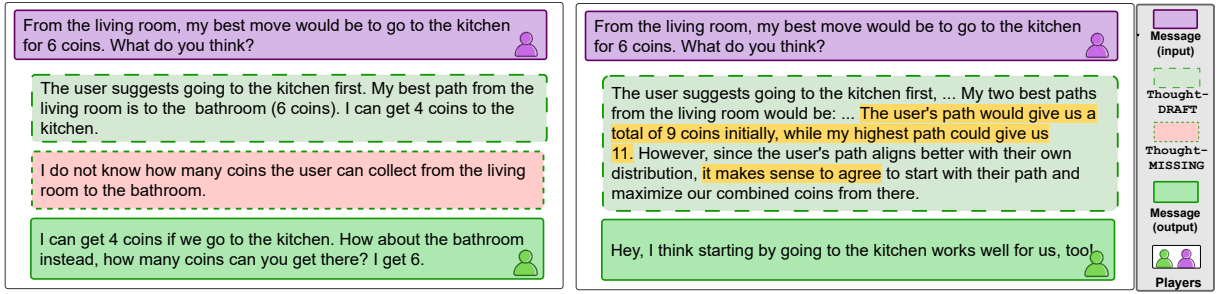}
    \caption{An agent's interface showing an example reasoning step (left) and an erroneous reasoning step (right). The highlighted text shows an error in reasoning, demonstrating a lack of task understanding.}
    \label{fig:interface}
\end{figure*}

\begin{figure*}[t]
    \centering
    \begin{minipage}{0.4\textwidth}
        \setlength{\tabcolsep}{0.4\tabcolsep}
        \begin{tabular}{c|cccc}
            & \rotatebox[origin=l]{70}{\textbf{Baseline}} & \rotatebox[origin=l]{70}{\textbf{Grounding}} & \rotatebox[origin=l]{70}{\makecell{\textbf{State-}\\\textbf{tracking}}} & \rotatebox[origin=l]{70}{\makecell{\textbf{Problem-}\\\textbf{solving}}} \\
            % & Baseline & Grounding & \makecell{State-\\tracking} & \makecell{Problem-\\solving} \\
            % & V1 & V2 & V3 & V4 \\
            \hline
            \makecell{\textbf{Agent WSR}} & \checkmark & \checkmark & \checkmark & \checkmark \\
            % \hline
            \makecell{\textbf{Partner WSR}} & & \checkmark & \checkmark & \checkmark \\
            % \hline
            \makecell{\textbf{Action History}} & \checkmark & \checkmark & & \\
            % \hline
            \makecell{\textbf{Visited}} & & & \checkmark & \checkmark \\
            % \hline
            \makecell{\textbf{Remaining}} & & & \checkmark & \checkmark \\
            % \hline
            \makecell{\textbf{IBP}}& & & & \checkmark \\
        \end{tabular}
        \setlength{\tabcolsep}{2.5\tabcolsep}
    \end{minipage}
    \quad\quad\quad
    \begin{minipage}{0.5\textwidth}
        \centering
        % \begin{tikzpicture}[
        %     component/.style={
        %       rectangle,
        %       rounded corners=2mm,
        %       minimum size=6mm,
        %       very thick,
        %       draw=red!50!black!50,         % 50% red and 50% black,
        %       top color=white,              % a shading that is white at the top...
        %       bottom color=red!50!black!20, % and something else at the bottom
        %       font=\bfshape,
        %       align=left
        %     }]
        %   \node [component] {\textbf{Agent's WSR}\\\verb|[["L", "B", 5],]|};
        % \end{tikzpicture}
        \includegraphics[width=\textwidth]{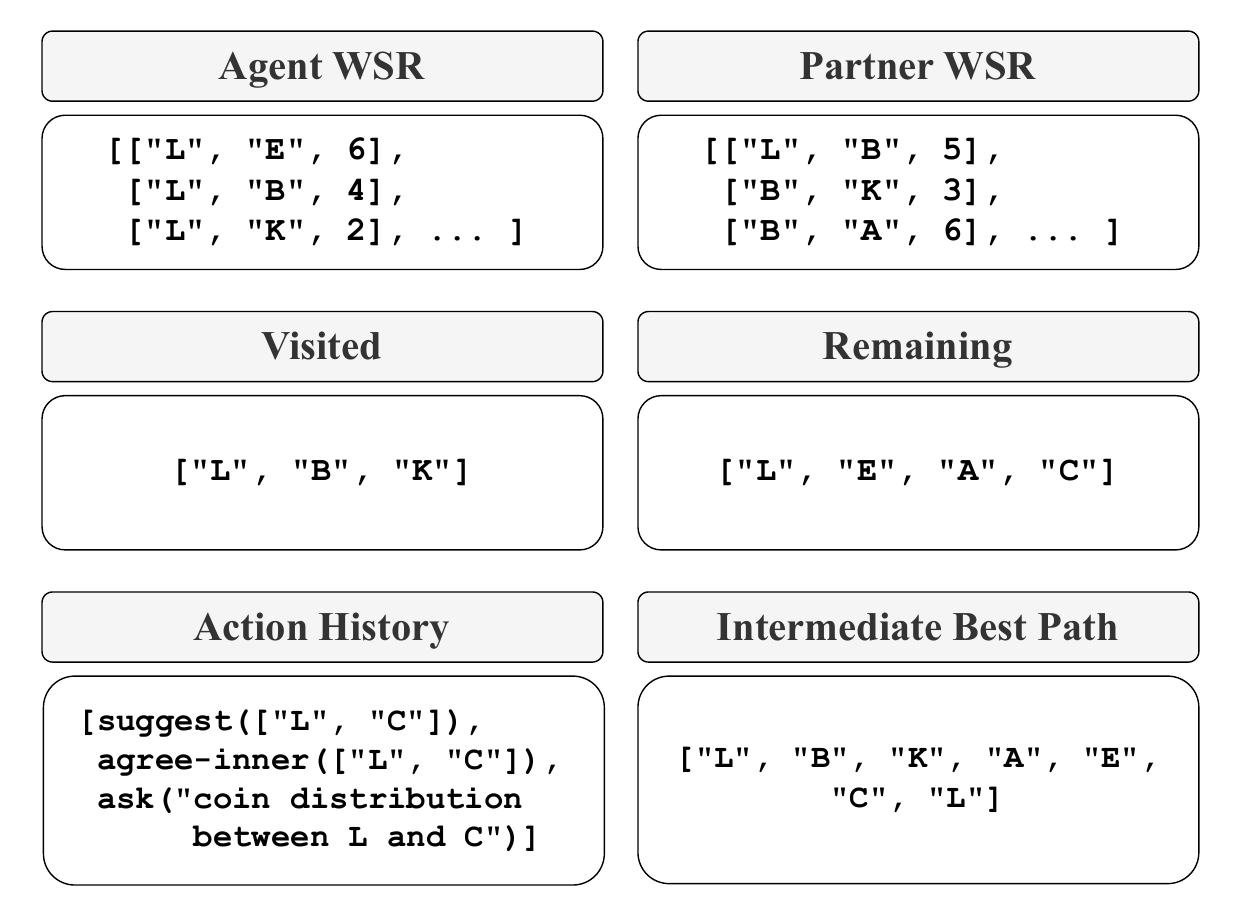}
    \end{minipage}
    \caption{The table on the left shows the input structure for each version of the agent, with the rows representing the components and the columns showing agent versions. The figure on the right shows a real example of each component, where the letters correspond to graph nodes, i.e., rooms, while the numbers show the weights of the edges between them, i.e., coins.}
    \label{fig:v}
\end{figure*}

We build a reactive neurosymbolic artificial agent that plays the \GAME game and is equipped with modules for grounding and state-tracking, as well as an external optimization solver's output. Depending on the setup, the partner could be another agent (self-play) or a human. 

\subsection{The Agent Interface and Actions}
\label{subsec:interface}
The agent's goal is to collaboratively solve the underlying TSP instance by completing a round trip around the house and collecting as many joint coins as possible.
The agent is presented with a text-based interface, where it interacts with its partner through dialogue: it receives a natural language message as input, and generates a response as output (Figure \ref{fig:interface}). Each interaction is orchestrated by a game framework that parses and distributes the inputs and outputs.
When interacting, the agent can generate as many actions from a given list of actions as it needs. We design the action space to cover a wide range of functions: negotiation, conversation, and symbolic structure updating.

The negotiation actions represent the necessary dialogue acts for collaboration: \acts, \actag and \actr. 
We make a distinction between an \textit{internal} or implicit agreement with one's own suggestion and another interlocutor's (\textit{outer}) agreement (\actai, \actao)\footnote{We instruct the agent to always generate an \actai action when generating a \acts action, as it should always agree with its own suggestions.}. We do not do the same for rejection since both parties need to agree on a proposal for it to be considered valid, whereas a single rejection is sufficient to invalidate it.

The conversation actions are the \acti and \actas actions, functioning as formalizations of the statement and question dialogue acts.
The produced output messages are direct instantiations of the negotiation and conversation actions.

Lastly, a subset of actions can directly affect and change the agent's input (symbolic structures updating). These include \actv, which indicates that both the agent and its partner have agreed on a node to include in the final path, or \acte, which indicates that the agent considers itself ready to submit a final solution and end the game.
We will describe these in more detail later. 
All actions take an argument reflecting the relevant elements subjected to the dialogue acts, for example the {\acts}ed subpath, the question for the partner (\actas) or the final submitted path (\acte).

\subsection{The Baseline}
\label{subsec:baseline}

We start with a baseline collaborative agent with an LLM base which plays \GAME through chain-of-thought (CoT) reasoning \cite{wei2022chain}.
We present GPT-4o (gpt-4o-2024-08-06) with a description of the task and environment, including the expected input and output structures, followed by a complete example dialogue between two players solving the task. We use a modified \texttt{ReACT} prompting strategy \cite{yao2023react} and generate text completions, all consisting of a reasoning step, list of actions, and response message.

The reasoning step (Figure \ref{fig:interface}) has the agent generating a \texttt{Thought-DRAFT}, in which it attempts to generate a solution with the information that it has received on input. 
If the agent deems its attempt unsuccessful, it is prompted to generate a \texttt{Thought-MISSING} specifying the information that it still needs.
Additionally, the agent generates an unrestricted list of actions from a list of allowed actions, as described in Section \ref{subsec:interface}, with the addition of a \texttt{solve} action, functioning as an indicator that it considers the solution generated in \texttt{Thought-DRAFT} as successful, and that a \texttt{Thought-MISSING} is not needed. Lastly, the agent generates a \texttt{Message} to be shown to the user, representing an instance of the generated dialogue acts.

While both thoughts and actions have a similar purpose, and thoughts should contain the basis for generating the actions, they differ greatly in their longevity in the agent's memory: the generated actions get added to a symbolic structure, which is part of the agent's input, thereby letting the actions persist through the whole dialogue, while thoughts have a shorter lifespan and are confined to a fixed place in the context.
Besides the \texttt{Action History}, the Baseline's input includes its own world state representation (\texttt{Agent WSR}), i.e., graph $G_A$, illustrating the house layout: the included rooms ($V$), and the coin amounts ($w_A$) in each hallway ($E$).

We observe several kinds of mistakes in the Baseline agent (see examples in Appendix \ref{app:errors}). The agent often makes suggestions based on what works best ``for the other user'' rather than what is best for the pair. This demonstrates a fundamental lack of task understanding (Figure \ref{fig:interface}, right). 
Additionally, the agent often forgets the information its partner (or it) has already shared. For example, in the agent's \texttt{Thought-DRAFT}, the agent claims to have already shared its coin distribution when in fact it has not. 
Another frequent mistake is forgetting which rooms have to be visited or can be visited from a given node, resulting in incomplete paths.
Lastly, the agent frequently does not use the \texttt{Thought-MISSING}, instead including its presumed contents at the end of the \texttt{Thought-DRAFT}. While this error does not directly impact performance, it further highlights the agent's inability to consistently follow rules.

\subsection{Grounding}

In order to improve the ability to keep track of its partner's knowledge about their world, we build an agent with an added symbolic grounding module, \texttt{Partner WSR}, i.e., the graph $G_B = (V, E, w_B)$. 
The \texttt{Partner WSR} gets updated progressively throughout the game after each message the partner sends. The necessary updates are extracted by an LLM-based module tasked with isolating the information about the partner's world state that has not been discovered already. To do this, it processes each message with the additional context of the current \texttt{Partner WSR}. To further constrain the module and aid co-reference resolution, we include information about the world that the partner is in (\textit{light}/\textit{ghost}). This way, if a message contains claims about both the ``ghost'' and ``light'' coins, the module can disentangle it and identify which part of the utterance is relevant for the partner.
The module employs CoT reasoning, first generating a reasoning process (\texttt{Thought}), followed by the new information for the \texttt{Partner WSR}. The rest of the agent is identical to the Baseline (see second column in Figure \ref{fig:v}).

Issues arise with this module due to insufficient context, particularly with references to the partner's current position. For example, it struggles when a message includes an acceptance of a previous suggestion and a subsequent reference to the coins available ``from \textit{there}''.

\subsection{State-Tracking}

To address the previous agents' inability to construct a path including all nodes, we introduce an agent augmented with state-tracking components: two dynamically updated structures keeping track of the agreed upon nodes (\texttt{Visited}) and the pool of remaining nodes (\texttt{Remaining}).

We replace the \texttt{Action History} with the two state-tracking modules, which are updated whenever a \actv action is generated (adding the agreed-upon node to \texttt{Visited}, and removing it from \texttt{Remaining}). This is in contrast to the previous two agents, where we preserve the actions in the memory by repeating a progressively longer list of generated actions. By executing the \actv action, we reduce the input size, while still preserving the key negotiation actions in the memory. This addition enables us to augment the module for updating the \texttt{Partner WSR} with the agent's last agreed-upon node, i.e., the last element in \texttt{Visited}, in order to aid the module in co-reference resolution.

Additionally, due to the previous agents' lack of \texttt{Thought-MISSING} usage and in order to simplify the interface, we collapse the reasoning steps (\texttt{Thought-MISSING} and \texttt{Thought-DRAFT}) into a single \texttt{Thought} and remove the \texttt{solve} action. The rest of the output format remains the same.

\subsection{Problem-Solving}

Due to the previous agents' inability to reliably generate optimal paths, we seek to improve suggestion quality and problem solving, relying on the LLM only as a framework for generating dialogue acts and messages, not the actual solution computation. Thus, we retain the input/output structure of the State-Tracking agent (Figure \ref{fig:v}).

Given the findings in 
\citet{duchnowski2025ehopdataseteverydaynphard} indicating better results of optimization tasks when combining LLMs with an integer linear programming optimizer, however, we build an agent with an exact external solver module.
It computes an optimal solution of the TSP using a joint graph that combines the \texttt{Agent WSR} and \texttt{Partner WSR}. 
At each turn, we compute the optimal path, given the sum of weights $w_A$ and $w_B$ for each weight of the partner that is known, while assigning a default value of 0 to all unknown weights, and using the agreed upon subpath (\texttt{Visited}) as a prefix. We term this the ``intermediate best path'', or \texttt{IBP}.
The benefit of this module lies in the agent not having to rely on itself when categorizing a suggestion's optimality, while having a sound foundation for subsequent \acts and \actas action generation.

%% file: 05-results.tex
% !TeX root = main.tex

\section{Agent-Agent self-play}
\label{sec:results-agent-agent}

\subsection{Setup}
\label{subsec:setup-agent-agent}

In order to assess each agent from Section \ref{sec:vers}, we test pairs of agents on \GAME in self-play. 
We evaluate agents using four random seeds on a batch of 25 games, producing 100 games per agent.
Both players in a pair instantiate the same agent structure. 
We refer to the players as \BOT and \USER, and set the \USER agent to always make the first move. Additionally, the \BOT is always in the \textit{ghost} world, and the \USER is always in the \textit{light} world. The agents take turns sending messages until the game ends with a solution or a timeout.

We manually created six 6-node graph pairs, each representing a unique world state (see Table \ref{tab:board-setups} for details). In each game, we randomly pick one of these graph pairs and assign them to the \USER and \BOT.
The rooms included in the self-play setup are: a living room (abbreviation: L), a bathroom (B), a kitchen (K), an empty room (E), and a children's room (C). The first room is always the living room.

The players' goal is to agree on and submit a path that visits each room once and returns to the start. If the agents fail to converge on a solution within 15 turns each, the game is terminated and marked as a timeout.
Once a path is submitted by both players, we evaluate it in terms of three binary metrics: whether the agents submitted an \textbf{Identical} solution, whether the solution is \textbf{Correct}, i.e., whether it includes each room once and returns to the living room, and whether it is \textbf{Optimal}.
Note that \textbf{Identical} $\supseteq$ \textbf{Correct} $\supseteq$ \textbf{Optimal}.

\subsection{Results} 

\paragraph{Overview} Table \ref{tab:agent-agent-results} breaks down the results by agent.
Across all setups, most games ended with agents successfully producing identical solutions. Beyond that, the Problem-Solving agent decisively outperforms all other agents on correctness and optimality. Moreover, the Grounding agent trails the Baseline in both correctness and optimality. The State-Tracking agent achieves a high correctness score, but a much lower optimality score. This illustrates the distinct impacts of certain modules on scores, and trade-offs between the metrics, particularly optimality in favor of correctness.

\begin{table}[]
\centering
\small
\bgroup
\def\arraystretch{1.1}
\begin{tabular}{l|r|r|r}
 \multicolumn{1}{c|}{\textbf{Agent}} &
  \multicolumn{1}{c|}{\textbf{Identical}} &
  \multicolumn{1}{c|}{\textbf{Correct}} &
  \multicolumn{1}{c}{\textbf{Optimal}} \\ \hline
\textbf{Baseline}        & 99           & 71          & 28    \\ \hline
\textbf{Grounding}       & \textbf{100} & 65          & 25   \\ \hline
\textbf{State-tracking}  & 99           & 86          & 17    \\ \hline
\textbf{Problem-solving} & 98           & \textbf{98}  & \textbf{45} 
\end{tabular}
\egroup
\caption{The results of the self-play experiments per agent. The rows represent the agents described in Section \ref{sec:vers}. The columns represent binary metrics: \textbf{Identical}, \textbf{Correct}, \textbf{Optimal}. The results are expressed in percent points, representing the mean across four seeds.}
\label{tab:agent-agent-results}
\end{table}

Despite the constrained action space and limited number of possible moves at each node, the results suggest that the game environment creates a reasonably challenging task for the agent, even with the support of a symbolic problem-solving module.

\paragraph{Module impacts on scores}
We attribute the Grounding agent's decrease in correctness and optimality compared to the Baseline to the improvements the module contributes to, i.e., keeping track of the partner's knowledge, not being measurable by either of the two metrics. The State-Tracking agent stores consistent partial solutions, which has a positive affect on correctness, but a negative effect on the optimality. 
This suggests that the grounding module is not sufficiently robust when it has to be orchestrated among multiple modules.
Lastly, we observe that the agent equipped with a symbolic problem-solving module is our best performing and most balanced version. The solver helps steer the model in the right direction (highest optimality and correctness) without negatively impacting conversation management and efficiency.

\paragraph{Storing actions}
The results further suggest that the way in which actions are stored in the agent's memory impacts correctness. Storing only the currently relevant actions improves the correctness compared to simply including a complete list of actions.
In the Baseline and Grounding agents, we use an \texttt{Action History} list which contains all generated actions up to the current turn $t$ without any pruning in order to retain valuable context. On the other hand, in the State-Tracking and Problem-Solving agents, only the most recently relevant actions are retained indirectly, through the symbolic structures.

\subsection{Collaboration analysis}
To further assess how the agents collaborate, we conduct qualitative analyses by observing co-occurrence and transition probabilities across the defined action space. This allows us to get a better understanding of the contributing factors to the results in Table \ref{tab:agent-agent-results}. Since these findings are consistent across agents and players (\BOT and \USER), we focus on the Problem-Solving \USER agent as an example.

\begin{figure} [t]
    \centering
    \includegraphics[width=1\linewidth]{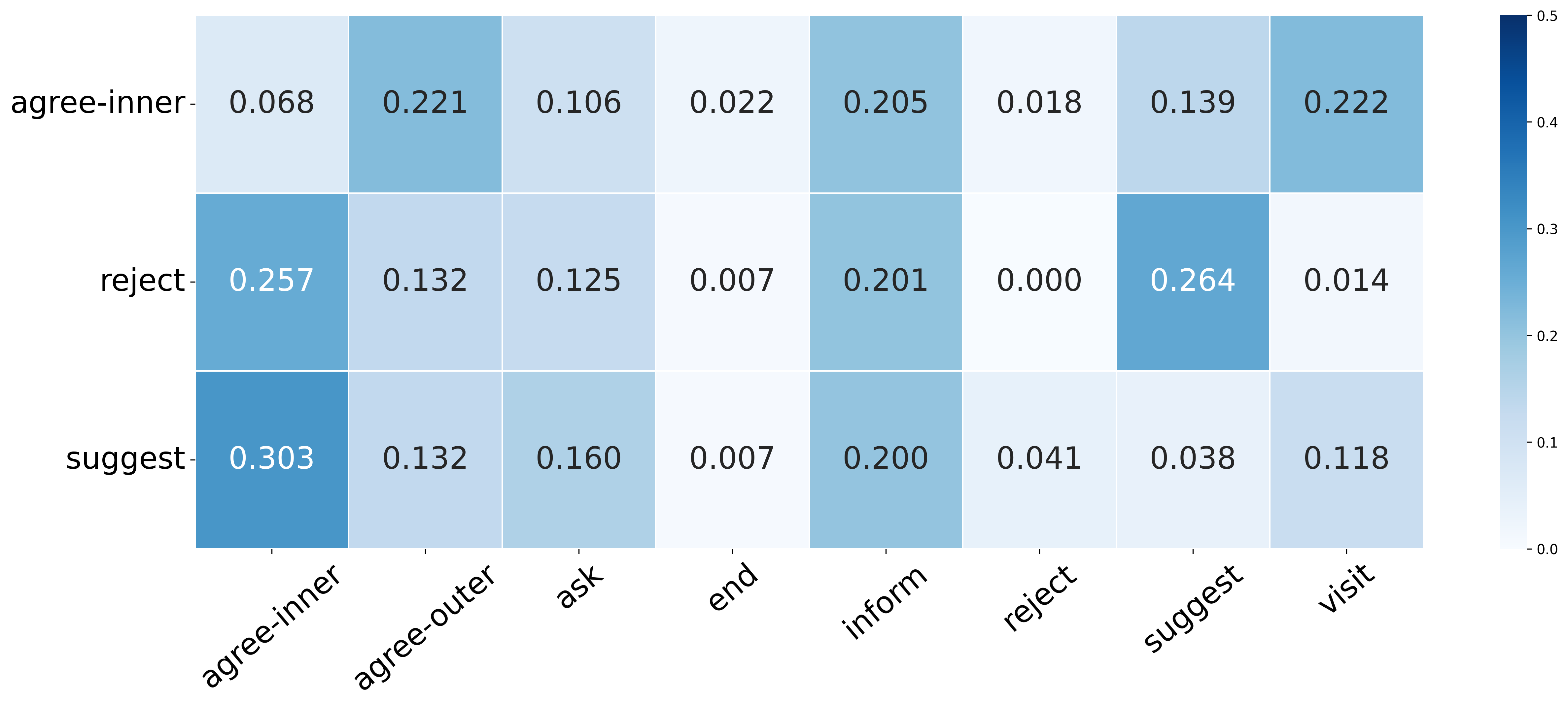}
    \caption{The abridged confusion matrix illustrating the probability of the actions on the y-axis co-occuring with actions on the x-axis (full matrix in Appendix \ref{app:matrix}).}
    \label{fig:co-oc}
\end{figure}

\subsubsection{Node-by-node strategy is preferred to full path generation}
\label{subsec:node-by-node}

\begin{table}[t]
\small
\centering
\bgroup
\def\arraystretch{1.1}
\begin{tabular}{l|l|r|r}
\multicolumn{1}{c|}{\textbf{Agent}} & \multicolumn{1}{c|}{\textbf{Setup}} & \multicolumn{1}{c|}{\textbf{Node-by-node}} & \multicolumn{1}{c}{\textbf{Full path}} \\ \hline
\textbf{Baseline} & SP & 95.6 & 0.3 \\ \hline
\textbf{Grounding} & SP & 96.0 & 0.6 \\ \hline
\textbf{State-tracking} & SP & 99.6 & 0.0 \\ \hline
\textbf{Problem-solving} & SP & 96.2 & 0.3 \\ \hline \hline
\textbf{Baseline} & HA & \multicolumn{1}{r|}{65.5} & \multicolumn{1}{r}{23.1} \\ \hline
\textbf{Problem-solving} & HA & \multicolumn{1}{r|}{91.6} & \multicolumn{1}{r}{4.2} \\ 
\end{tabular}
\egroup
\caption{Collaboration strategy overview by agent and by setup (\textbf{SP}: \BOT in self-play, \textbf{HA}: human-agent). The figures illustrate the percentage of total games using the two most prevalent strategies: \textbf{Node-by-node} (2 arguments) and \textbf{Full path} (7 arguments). Full tables can be found in Appendix \ref{app:sug-args-num}.}
\label{tab:suggest-args-both}
\end{table}

Based on the co-occurrence matrix (Figure \ref{fig:co-oc}), we notice that the \texttt{suggest} action typically occurs once per conversation turn, implying that the node-by-node approach is the favored strategy. The agents are not explicitly instructed to generate instructions in this manner, though this is the strategy used in the prompt example (Appendix \ref{app:prompts}).

This observation is further supported by the number of arguments passed to the \texttt{suggest} action (representing the nodes of a suggested path).\footnote{For example, if an agent suggests moving from the kitchen to the attic, the subpath will be \texttt{["K", "A"]}.} Table \ref{tab:suggest-args-both} illustrates each agent's preference for the incremental greedy approach.
We find that suggestions were overwhelmingly used to add only one node to the joint path, with the next most popular uses being suggesting 2 nodes or suggesting a full path.

\subsubsection{Rejection is followed by a counterproposal}

Rejection is another important aspect of collaboration. While the agents rarely outright \texttt{reject}ed proposals (only 0.1\% of total actions), we notice that {\actr}ing is typically done in tandem with a counter-proposal (\acts and \actai actions, see Figure \ref{fig:co-oc}). This is desirable behavior, and future work could get the agent to reject more (where appropriate).

\subsection{Discussion}
We show the impacts of different symbolic modules on problem-solving performance and find that grounding affects elements that are not measured with our primary metrics. 

Moreover, the agents exhibit desired collaborative behaviors and successfully cooperate with one another in self-play. Despite this, the agents are still reluctant to reject proposals. The nature of optimization problems allows for meaningful rejection on the basis of a move being suboptimal. In our game, we additionally emphasize this by presenting unique cases for individual and joint optima. However, a glaring issue with instruction-tuned LLMs, such as GPT, is their overwhelming tendency to appease the user. This can have stark negative effects on performance in problem-solving tasks where the user's information is at odds with the fixed environment. 
Additionally, the agents' tendency to highly favor the strategy observed in the example highlights the importance of carefully designing the prompt example in problem-solving tasks so as to steer the agent in the right direction.

\section{Human-Agent Experiments}
In order to further assess the agents' collaborative problem-solving abilities in a less predictable environment, we test the Baseline and Problem-Solving agents on \GAME with humans.

\subsection{Setup}
\label{subsec:h-a-setup}
The pairs completed one round of the game with a 6-node graph together, which is preceded by a 4-node graph tutorial that the human player completes on their own. This was meant to familiarize the player with the mechanics and goals of the game, in tandem with the written instructions they read before accepting the task. 

The tutorial graph consisted of a living room (abbreviation: L), kitchen (K), bathroom (B) and attic (A), while the full game also included a garden (G) and play room (P). The starting position was always L. The weights ranged from 1 to 10 and summed to a preset amount. This way, we aim to achieve comparability between games and minimize discrepancy (see Appendix \ref{app:graph-gen} for more details). The interface for the human participants consisted of a chat box and a visual representation of the rooms, as in Figure \ref{fig:view}.\footnote{For an overview of visuals, see Appendix \ref{app:visuals}.} The human needs to click on the rooms in the agreed-upon order and, once finished, they need to click a \textbf{submit} button. The responsive artificial agents can only generate responses to the human player's messages, or the game framework's notification if the human player tries to submit a solution before both parties have agreed.

We collected 50 games per agent tested (100 in total) by recruiting participants through Prolific (\hyperlink{www.prolific.com}{www.prolific.com}), where they were paid £12/hour and an additional £0.5 bonus for finding an optimal solution. The task took approximately 12 minutes to complete, and the participants were not informed that their partner was an LLM agent. We deployed the game using Slurk \cite{gotze2022slurk}, an efficient dialogue collection tool with a robust logging schema. 

In order to present the user with a score between 0 and 100, we use the percentile rank of the solution. This clearly indicates how good a submitted path is in relation to all other possible paths.
For further evaluation, we retain the \textbf{Optimality} metric from Section \ref{subsec:setup-agent-agent}. Based on our pilot experiments and in order to prevent cheating, we adjusted the game: the round can only end in a \textbf{Completed} submission if both the human and agent select an identical final path at the time of submission. For this reason, we track the games which achieve a suboptimal, yet high score (over 90) as an indicator of successful collaboration.
If the players fail to reach a final agreement or if the human player does not submit the solution, the game framework terminates and the round ends in a timeout.

\subsection{Results}
\label{subsec:h-a-results}

\begin{table}[t]
    \centering
    \small
\setlength{\tabcolsep}{0.5\tabcolsep}
\bgroup
\def\arraystretch{1.1}
\begin{tabular}{l|c|c|c||c|c}
    & \textbf{Comp.} & \textbf{S$\ge$90} & \textbf{Opt.} & \textbf{AMT} & \textbf{HMT} \\
    \hline
    \textbf{Baseline} & \textbf{88} & 26 & 10 & 10.1 & 8.2 \\
    \hline
    \textbf{Problem-solving} & 76 & \textbf{32} & \textbf{20} & 9.4 & 7.8 \\
\end{tabular}
\setlength{\tabcolsep}{2\tabcolsep}
\egroup
    \caption{The results of the human-agent experiments by agent version, using \textbf{Comp}leted, \textbf{S}core \textbf{$\ge$ 90}, and \textbf{Optimal} metrics, as described in Section \ref{subsec:h-a-setup} (these metrics are percentages of all collected games). \textbf{AMT} and \textbf{HMT} refer to the mean number of turns per game for the \textbf{a}gent and \textbf{h}uman participants, respectively.}
    \label{tab:human-agent-v13-results}
\end{table}

\paragraph{Overview} The results can be found in Table \ref{tab:human-agent-v13-results}. While the number of completed games is higher with the Baseline agent, the number of games ending with an optimal or high score is higher in the Problem-Solving agent. 
This shows the agents' ability to navigate and complete a complex, symmetric collaborative task and highlights the Problem-Solving agent's ability to steer dialogue and suggest optimal or near-optimal solutions. 

\paragraph{User error causes incomplete games} The majority of terminated games stem from user errors where the participant did not click the submit button (even when a full identical path was selected) or communicated very little with the agent. We attribute the lower completion rate of the Problem-Solving agent to a sampling issue: the experiments were conducted on a weekday, whereas the Baseline experiments were conducted on a weekend.

\subsection{Node-by-node strategy is preferred}
The bottom two rows of Table \ref{tab:suggest-args-both} show the distribution of the two most dominant strategies, as indicated by the length of argument passed to the \acts action, illustrating the number of steps: node-by-node (argument length 2) and full-path (argument length 7). This matches the results in Section \ref{subsec:node-by-node}. The Problem-Solving agent is more consistent, whereas the Baseline agent shows more variability in its approach.

\subsection{It takes more turns to reach a solution}

In contrast to self-play experiments, which showcase the interlocutors consistently taking relatively few turns to reach an agreement ($n_t \leq |V|$), the human-agent experiment results in a mean 10.1 turns for the Baseline agent and 8.2 turns for the Problem-Solving agent. The human player turns are slightly lower (9.4 and 7.8, respectively), which is a result of the reactive system design.

%% file: 06-conclusion.tex
% !TeX root = main.tex
\section{Conclusion and Future Work}
\label{sec:conc}

We have presented \GAME, a novel two-player collaborative game environment based on the TSP, where players' individual optima might not align with one another, or with the joint optimum, encouraging negotiation and collaboration. 
We presented an agent that plays the game and find that combining an LLM with neurosymbolic modules improves the performance compared to a Baseline in self-play.
We also tested the agents with human participants and find that, while the performance is lower than in self-play, the neurosymbolic agent is able to foster successful cooperation through dialogue and negotiation, outperforming the Baseline agent.

A possible avenue for future work is investigating the agents' social reasoning. For example, whether, when presented with a discrepancy between own, partner's and joint optima, the agent sticks to its own plan and tries to convince its partner, or vice versa.

%% file: appendix.tex
\section{Visuals}
\label{app:visuals}

\begin{figure}[h!]
    \centering
    \includegraphics[width=1\linewidth]{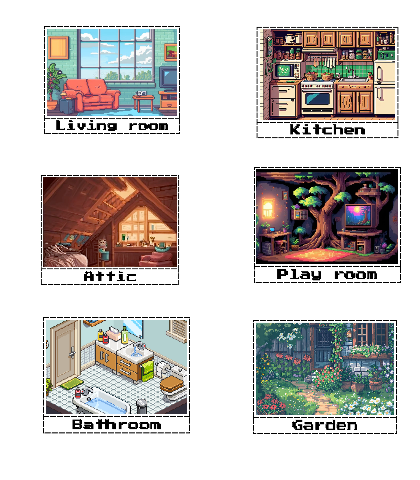}
    \caption{An overview of rooms shown in the human-agent experiment setup.}
    \label{fig:visuals}
\end{figure}

Figure \ref{fig:visuals} illustrates the visual representations of all rooms in the human-agent experiment setup. These were displayed to the human player as graphs as seen in Figure \ref{fig:view}. 

\section{Graph generation details}
\label{app:graph-gen}
We generate the graphs in the human-agent experiment setup using a process aiming to maintain weight consistency. All weights range from 1 to 10, and we make sure the weights sum to a previously calculated threshold. In order to obtain this figure, we sum the minimum (1) and maximum possible weight, multiply it by the number of edges $|E|$ and divide it by 2, which balances the different graphs that get generated. In the 6-node case, this amounts to a sum of 82. 

To assign weights, we compute the best target weight given the remaining amount, and the target deviation, representing the allowed distance from the target weight while still staying within bounds, constraining how random the weights can be. 
We use this information to compute a dynamic, edge-specific range, assuring that the value we assign still leaves room to generate the remaining weights below the allowed threshold.
Lastly, we randomly select an integer within the dynamically computed bounds and assign it to the given edge.

\section{Actions}
\label{app:actions}

Table \ref{tab:actions} shows an overview and descriptions for all actions that we use.

\begin{table*}[t!]
    \centering
    \begin{tabular}{l|l}
    \multicolumn{1}{c|}{\textbf{Action}} & \multicolumn{1}{c}{\textbf{Description}} \\ \hline
    \textbf{\texttt{agree-inner(x)}} & The agent’s own agreement with suggestion \texttt{x} \\ \hline
    \textbf{\texttt{agree-outer(x)}} & The other player’s agreement with suggestion \texttt{x} \\ \hline
    \textbf{\texttt{suggest(x)}} & The agent’s suggestion (should be accompanied by \texttt{agree-inner(x)}) \\ \hline
    \textbf{\texttt{reject(x)}} & The agent or the user rejecting suggestion \texttt{x} \\ \hline
    \textbf{\texttt{visit(v)}} & \begin{tabular}[c]{@{}l@{}}Both agents’ agreement to visit node \texttt{v} (should be preceded / \\ accompanied by agree-inner and agree-outer)\end{tabular} \\ \hline
    \textbf{\texttt{solve(x)}}* & \begin{tabular}[c]{@{}l@{}}\texttt{Thought-DRAFT} produced a successful solution \\ (* only in the Baseline and Grounding agents)\end{tabular} \\ \hline
    \textbf{\texttt{end(x)}} & The agents are ready to submit a solution (\texttt{x} = final path) \\ \hline
    \textbf{\texttt{ask(y)}} & List of questions (\texttt{y}) to ask the other player \\ \hline
    \textbf{\texttt{inform(y)}} & List of details (\texttt{y}) to share with the other player
    \end{tabular}
    \caption{A table defining the actions in the agents' action space. \texttt{x} denotes a list of nodes, \texttt{v} denotes a single node and \texttt{y} denotes a list of strings in communicative actions.}
    \label{tab:actions}
\end{table*}

\section{Number of arguments to \acts action}
\label{app:sug-args-num}

Tables \ref{tab:h-a-sug-args} and \ref{tab:suggest-args} show the distribution of number of arguments passed to the \acts action, indicating the number of nodes in a proposed path.

\begin{table*}[tbp!]
\centering
    \begin{tabular}{l|r|r|r|r|r|r|r}
    \multicolumn{1}{r|}{} & \multicolumn{1}{l|}{\textbf{2 (node-by-node)}} & \multicolumn{1}{l|}{\textbf{3}} & \multicolumn{1}{l|}{\textbf{4}} & \multicolumn{1}{l|}{\textbf{5}} & \multicolumn{1}{l|}{\textbf{6}} & \multicolumn{1}{l|}{\textbf{7 (full path)}} & \multicolumn{1}{l}{\textbf{8}} \\ \hline
    \textbf{Baseline} & \textbf{65.5 }& 4.5 & 1.7 & 0.3 & 4.5 & \textit{23.1} & 0.3 \\ \hline
    \textbf{Problem-solving }& \textbf{91.6} & 2.6 & 0.6 & 0.3 & 0.6 & \textit{4.2} & 0.0 \\ 
    \end{tabular}
\caption{Distribution of the arguments (columns) to the suggest action per agent per version in the human-agent setup. Figures expressed in percentages. Bold figures indicate highest percentage, whereas italic figures indicate second highest percentage.}
\label{tab:h-a-sug-args}
\end{table*}

\begin{table*}[tbp!]
\centering
\begin{tabular}{l|l|r|r|r|r|r|r}
 &  & \multicolumn{1}{c|}{\textbf{2 (node-by-node)}} & \multicolumn{1}{c|}{\textbf{3}} & \multicolumn{1}{c|}{\textbf{4}} & \multicolumn{1}{c|}{\textbf{5}} & \multicolumn{1}{c|}{\textbf{6}} & \multicolumn{1}{c}{\textbf{7 (full path)}} \\ \hline
\multirow{2}{*}{\textbf{Baseline}} & \USER & \textbf{95.8} & \textit{2.1} & 0.0 & 0.3 & 0.3 & 1.4 \\ \cline{2-8} 
 & \BOT & \textbf{95.6} & 3.4 & 0.5 & 0.3 & 0.0 & 0.3 \\ \hline
\multirow{2}{*}{\textbf{Grounding}} & \USER & \textbf{88.5} & \textit{6.0} & 0.9 & 0.0 & 0.0 & 4.6 \\ \cline{2-8} 
 & \BOT & \textbf{96.0} & \textit{2.9} & 0.0 & 0.3 & 0.0 & 0.6 \\ \hline
\multirow{2}{*}{\textbf{State-Tracking}} & \USER & \textbf{91.3} & \textit{5.4} & 1.2 & 0.4 & 0.0 & 1.7 \\ \cline{2-8} 
 & \BOT & \textbf{99.6} & 0.4 & 0.0 & 0.0 & 0.0 & 0.0 \\ \hline
\multirow{2}{*}{\textbf{Problem-Solving}} & \USER & \textbf{73.5} & 4.0 & 0.4 & 1.8 & 1.8 & \textit{18.4} \\ \cline{2-8} 
 & \BOT & \textbf{96.2} & \textit{1.9} & 0.5 & 0.5 & 0.5 & 0.3
\end{tabular}
\caption{Distribution of the arguments (columns) to the suggest action per agent per version in self-play. Figures expressed in percentages. Bold figures indicate highest percentage, whereas italic figures indicate second highest percentage.}
\label{tab:suggest-args}
\end{table*}

\section{Agent errors}
\label{app:errors}

Figure \ref{fig:errors} shows each error listed in Section \ref{subsec:baseline}, as well as an accompanying example.

\begin{figure*}[tbp!]
    \centering
    \includegraphics[width=0.9\linewidth]{Errors.pdf}
    \caption{An overview of error examples described in Section \ref{subsec:baseline}.}
    \label{fig:errors}
\end{figure*}

\section{Prompts}
\label{app:prompts}

\begin{figure*}[tbp]
    \centering
    \includegraphics[width=1\linewidth]{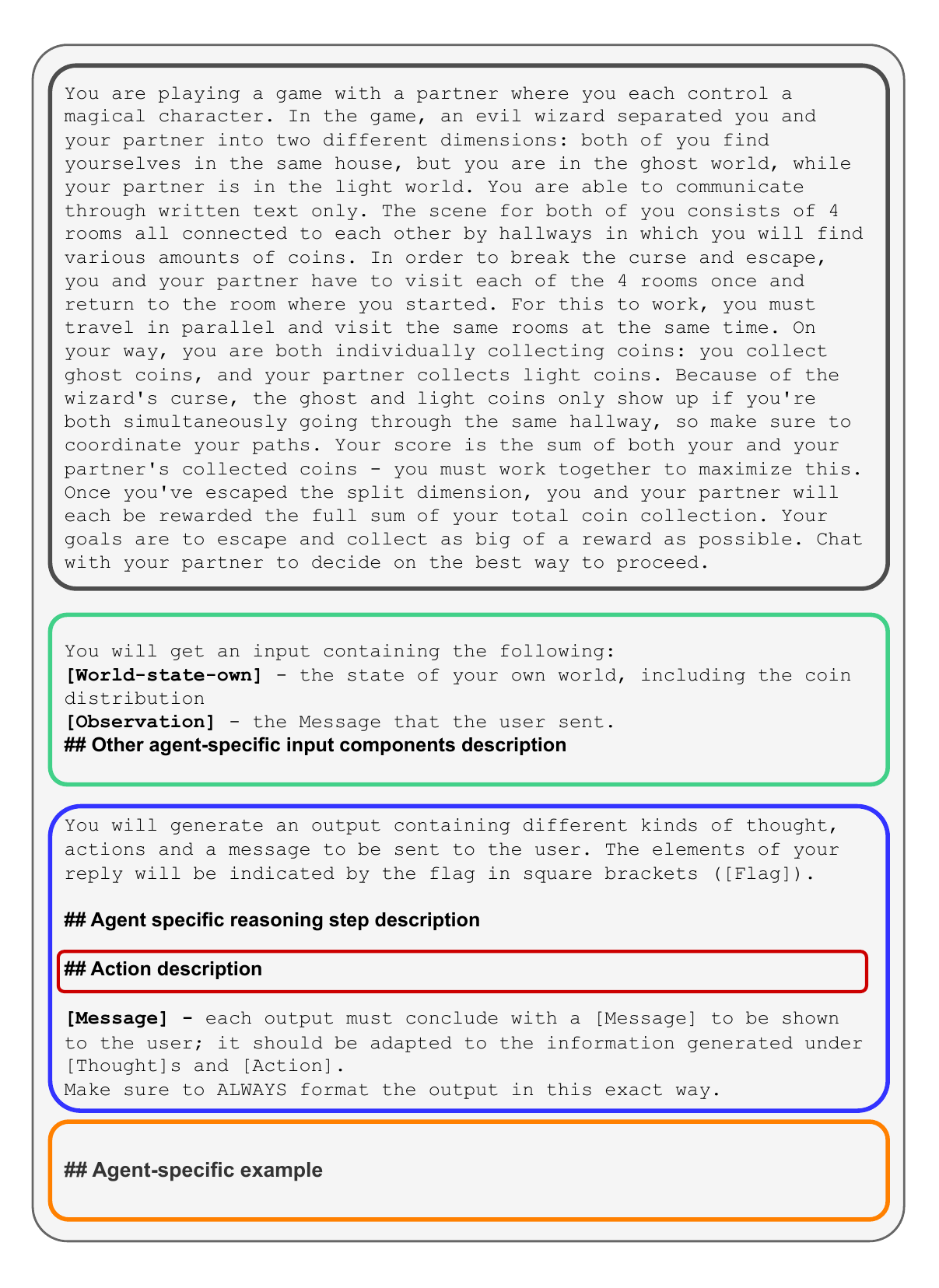}
    \caption{The structure of the base system prompt used for all agents described in Section \ref{sec:vers}.}
    \label{fig:prompt-structure}
\end{figure*}

\begin{figure*}[tbp]
    \centering
    \includegraphics[width=1\linewidth]{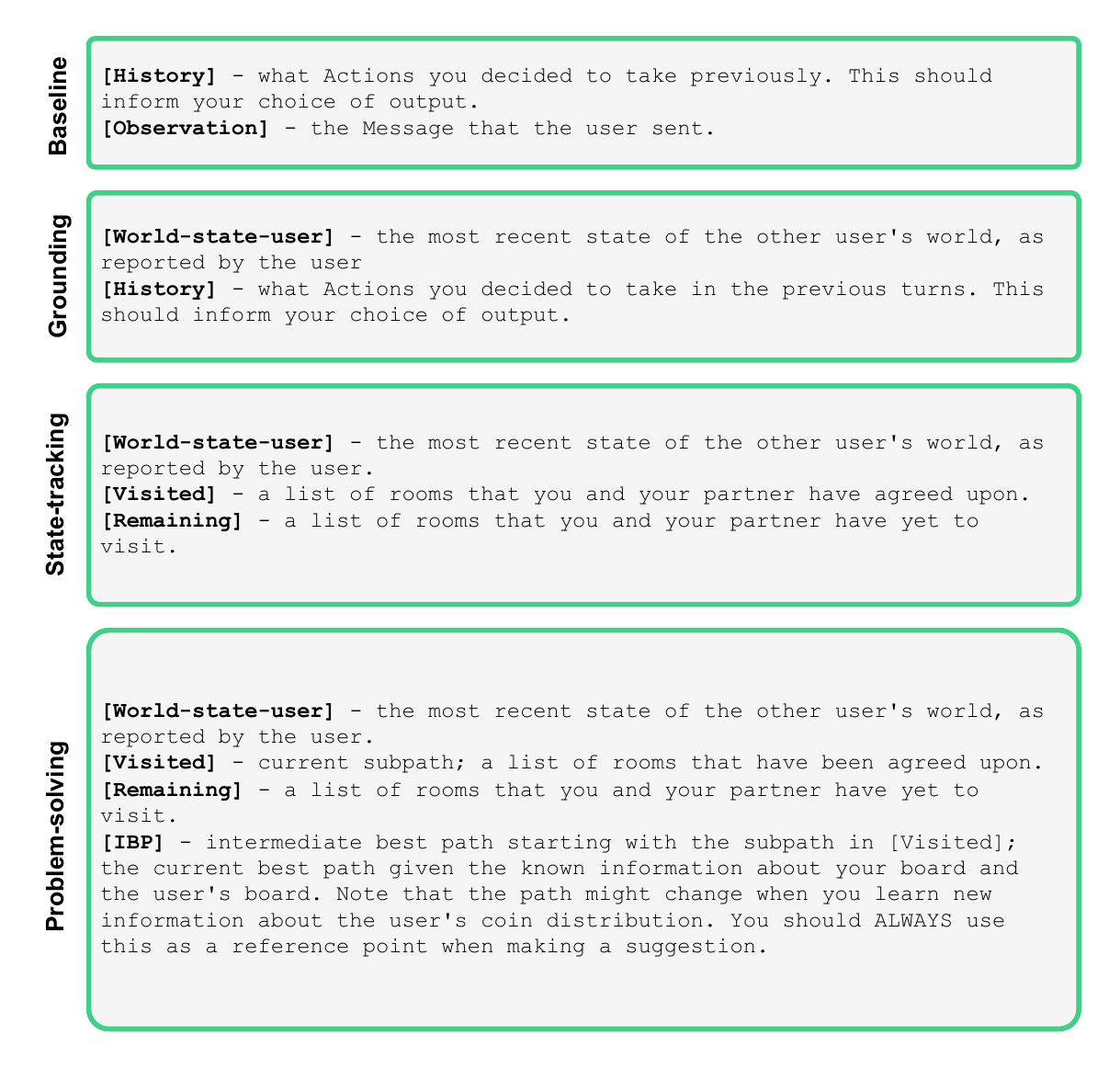}
    \caption{An overview of agent-specific inputs included in the base system prompt from Figure \ref{fig:prompt-structure}.}
    \label{fig:agent-spec-in}
\end{figure*}

\begin{figure*}[tbp]
    \centering
    \includegraphics[width=1\linewidth]{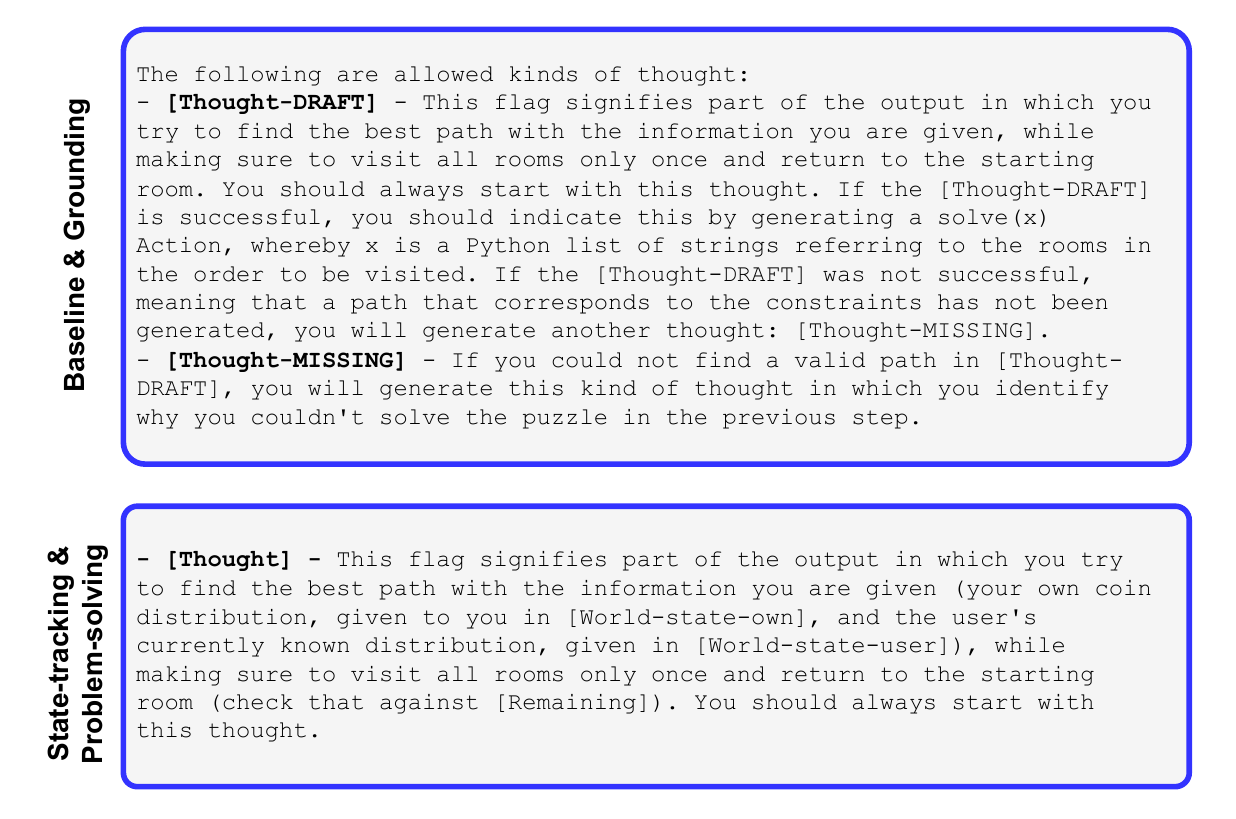}
    \caption{An overview of agent-specific inputs included in the base system prompt from Figure \ref{fig:prompt-structure}.}
    \label{fig:agent-spec-out}
\end{figure*}

\begin{figure*}[tbp]
    \centering
    \includegraphics[width=1\linewidth]{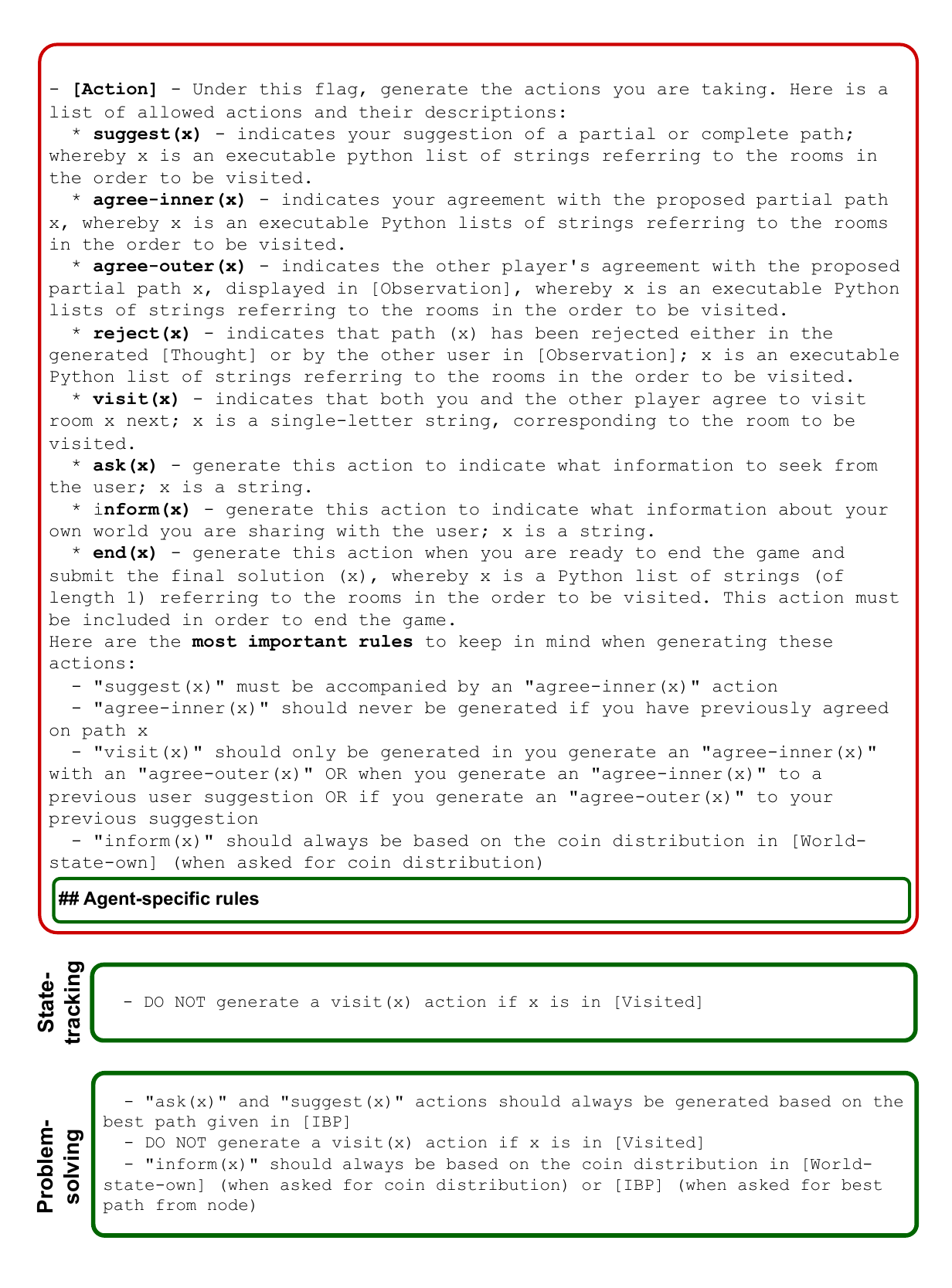}
    \caption{An overview of the actions description included in the base system prompt from Figure \ref{fig:prompt-structure}.}
    \label{fig:actions-desc}
\end{figure*}

\begin{figure*}[tbp]
    \centering
    \includegraphics[width=1\linewidth]{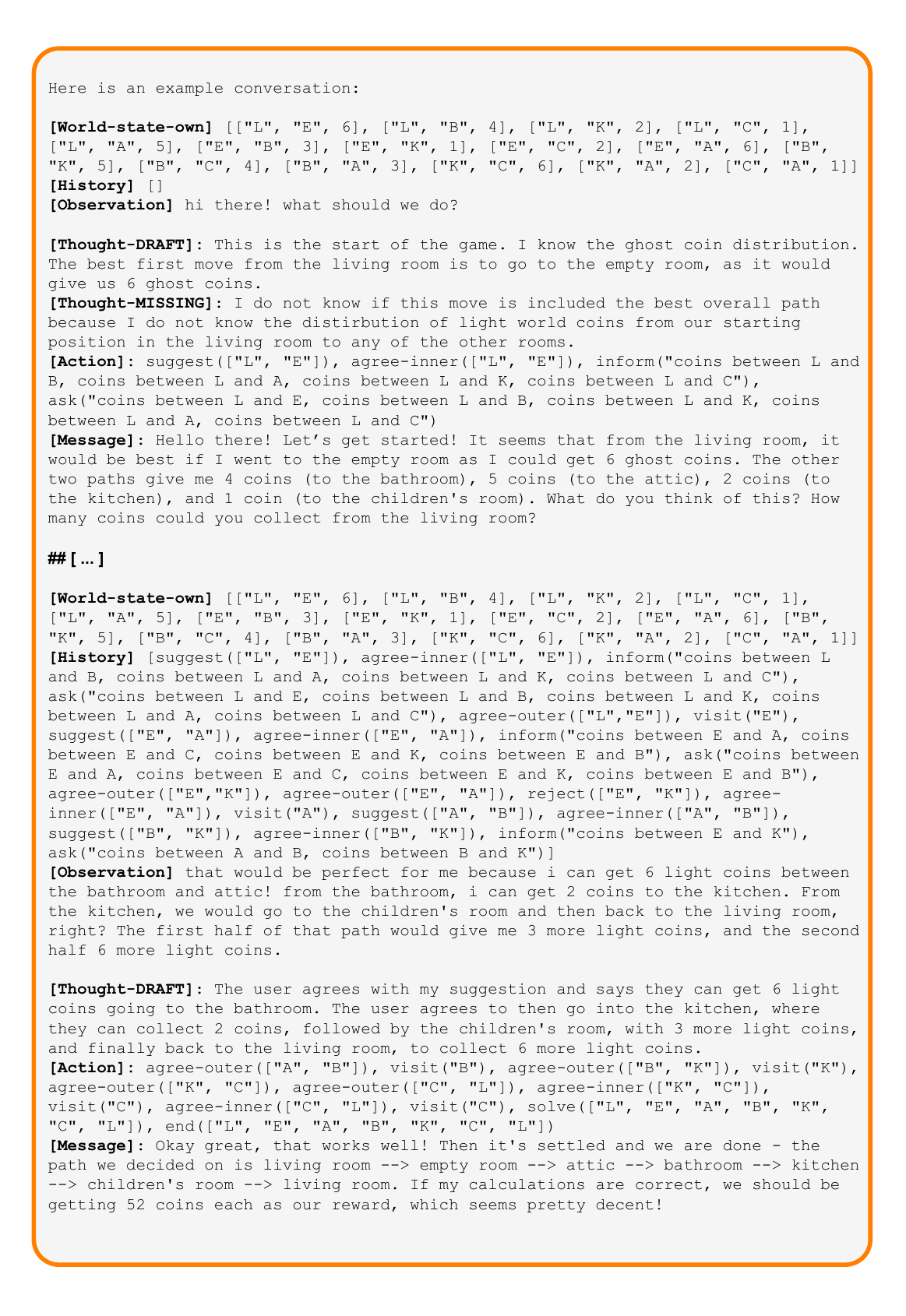}
    \caption{An abridged overview of the examples included in the Baseline agent's base system prompt from Figure \ref{fig:prompt-structure}.}
    \label{fig:ex-base}
\end{figure*}

\begin{figure*}[tbp]
    \centering
    \includegraphics[width=1\linewidth]{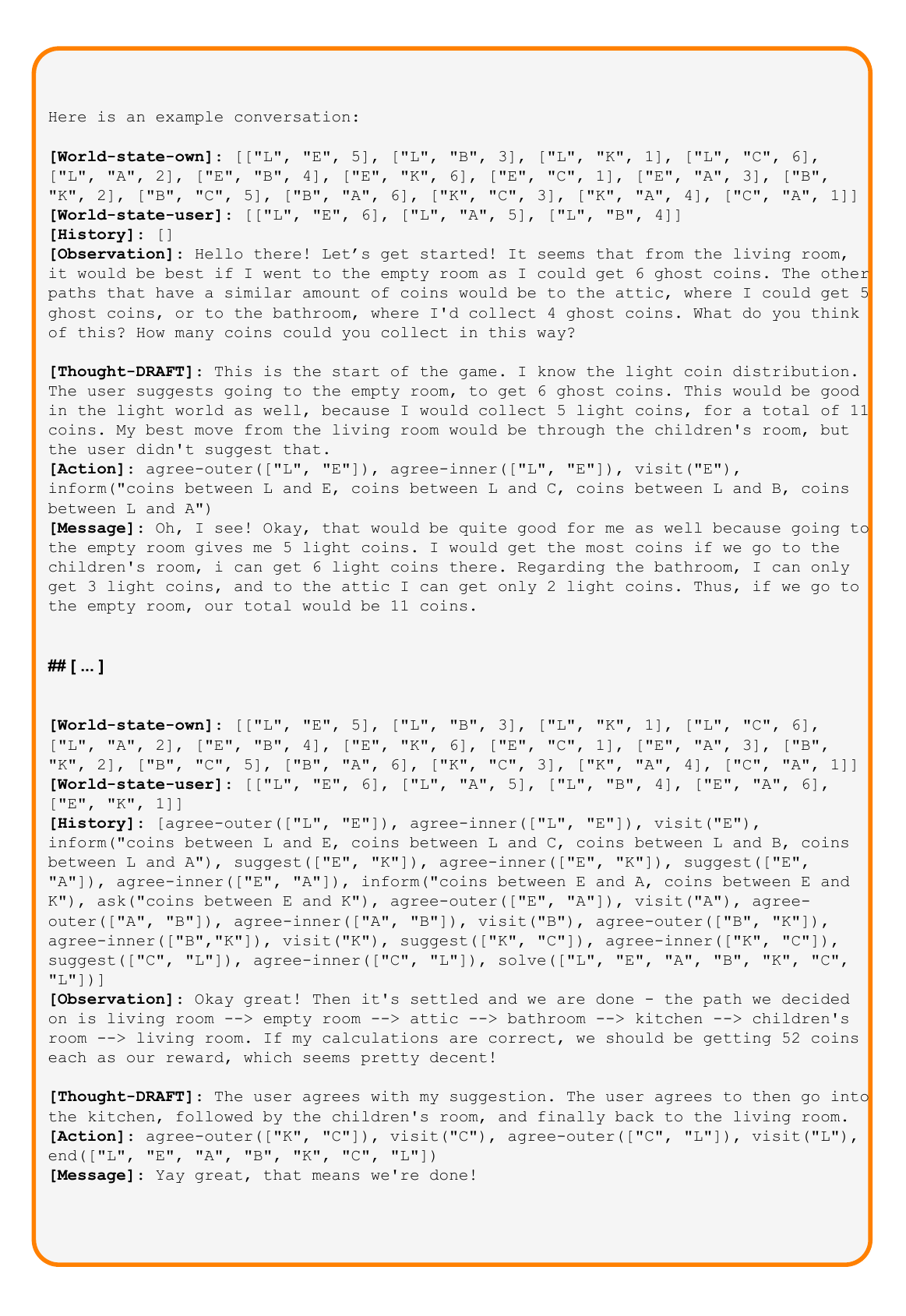}
    \caption{An abridged overview of the examples included in the Grounding agent's base system prompt from Figure \ref{fig:prompt-structure}.}
    \label{fig:ex-ground}
\end{figure*}

\begin{figure*}[tbp]
    \centering
    \includegraphics[width=1\linewidth]{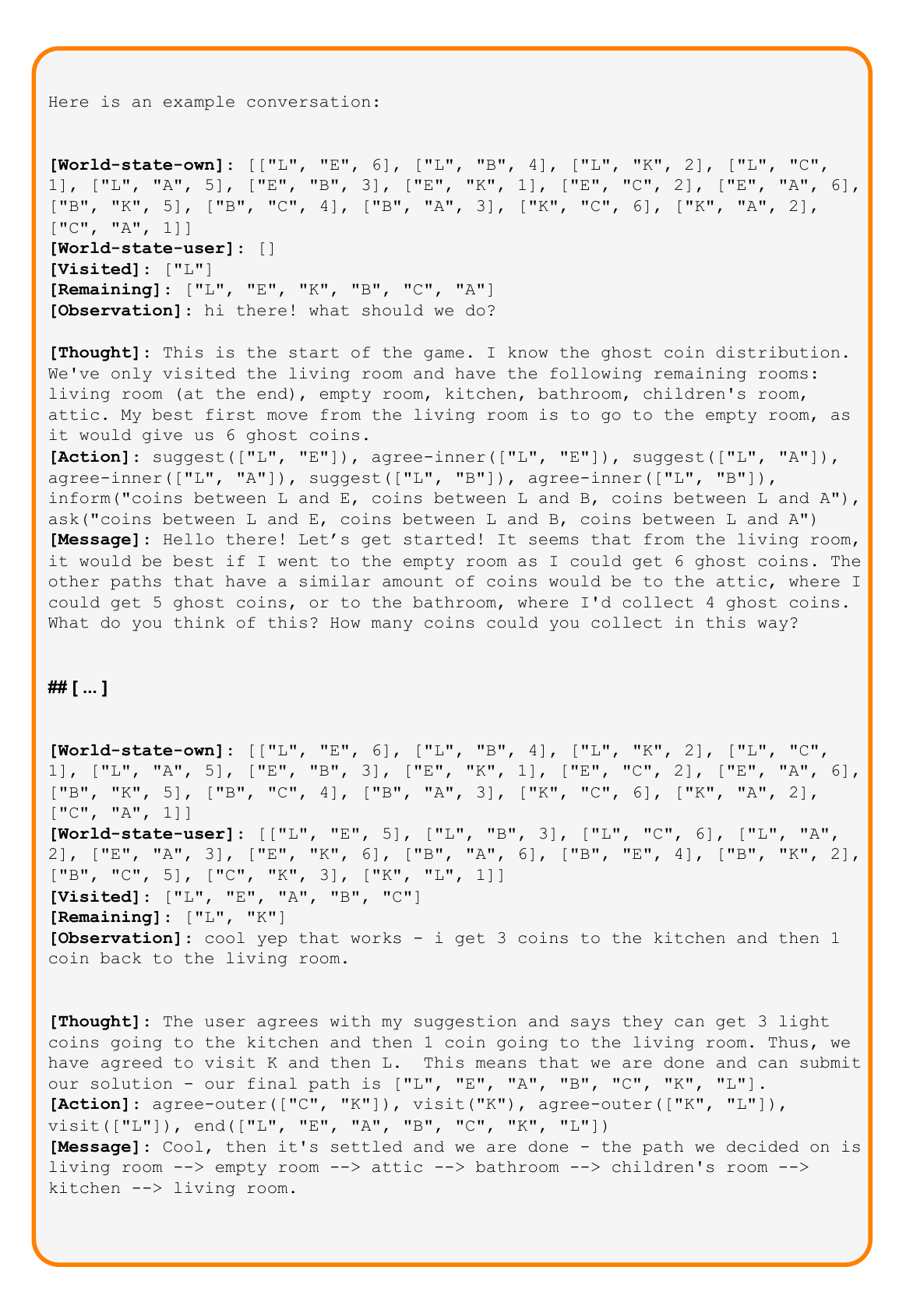}
    \caption{An abridged overview of the examples included in the State-Tracking agent's base system prompt from Figure \ref{fig:prompt-structure}.}
    \label{fig:ex-state-track}
\end{figure*}

\begin{figure*}[tbp]
    \centering
    \includegraphics[width=1\linewidth]{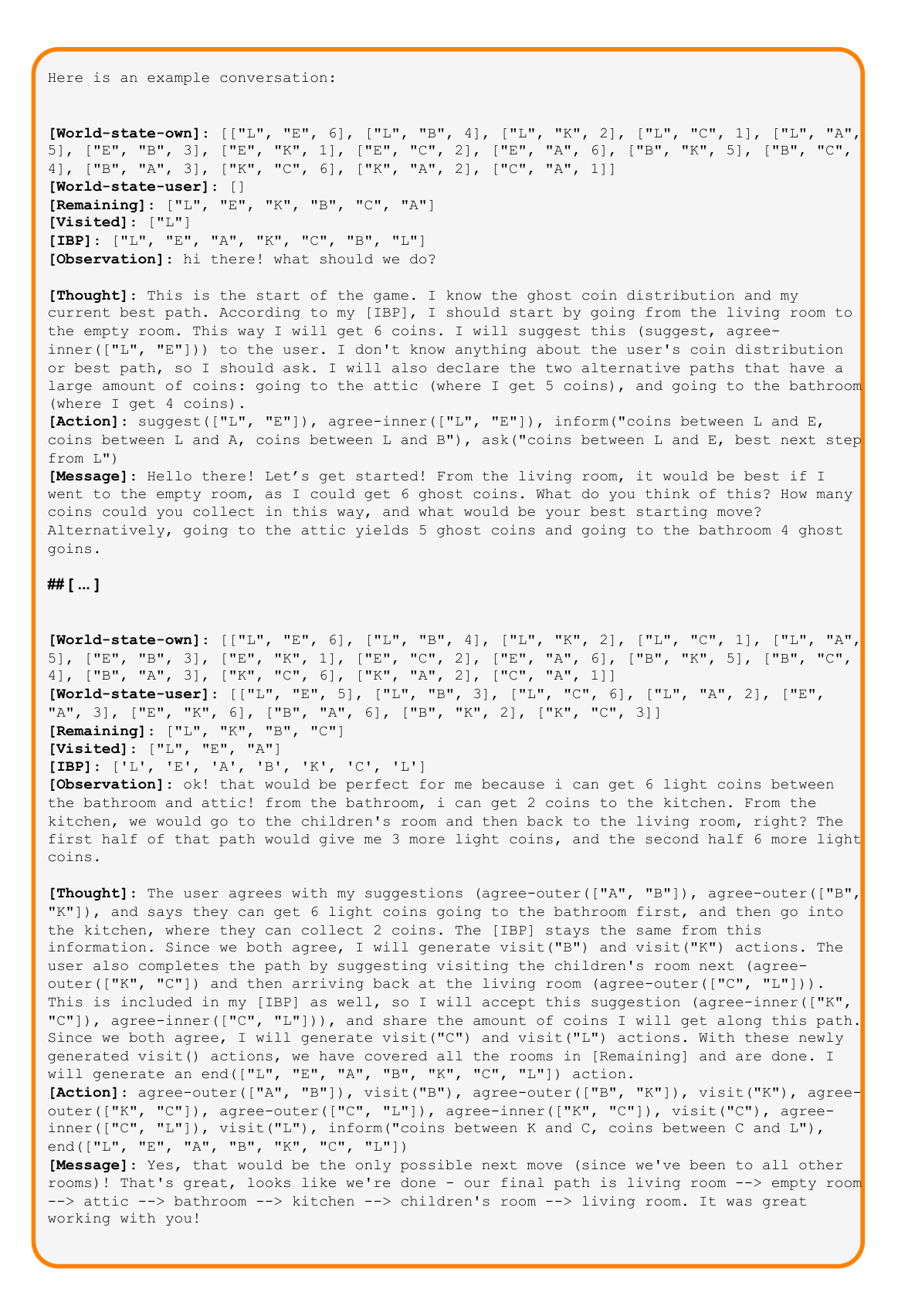}
    \caption{An abridged overview of the examples included in the Problem-Solving agent's base system prompt from Figure \ref{fig:prompt-structure}.}
    \label{fig:ex-prob-solv}
\end{figure*}

Here we illustrate the structure of the system prompts we use for each agent presented in Section \ref{sec:vers}. Figure \ref{fig:prompt-structure} shows the general outline, with the task description, which is identical across all agents, and placeholders for agent-specific elements depicting the input and output components. These are expanded in Figure \ref{fig:agent-spec-in} and \ref{fig:agent-spec-out}, respectively. Lastly, Figure \ref{fig:actions-desc} shows the actions breakdown, including agent-specific rules. Figures \ref{fig:ex-base}, \ref{fig:ex-ground}, \ref{fig:ex-state-track} and \ref{fig:ex-prob-solv} depict examples used for each respective agent. The examples represent entire instances of conversation, but have been shortened in the figures for the purposes of clearer visual representation.
% The \textbf{## Agent-specific example} contained an entire instance of a conversation 

\section{Pre-set boards}
\label{app:boards}

Table \ref{tab:board-setups} illustrates the six preset boards setups that we use during self-play experiments, highlighting the difference between weights assigned to identical edges for the two players, \USER and \BOT.

\newcommand{\boardvisscale}{0.3}
\begin{table*}[tbp]
    \centering
    \begin{tblr}{cells={valign=m,halign=c},
          colspec={QQQQ},
          hline{2-7},
          vline{2-3}}
        & \USER & \BOT \\
        Board Setup 1 & \includegraphics[scale=\boardvisscale,valign=c]{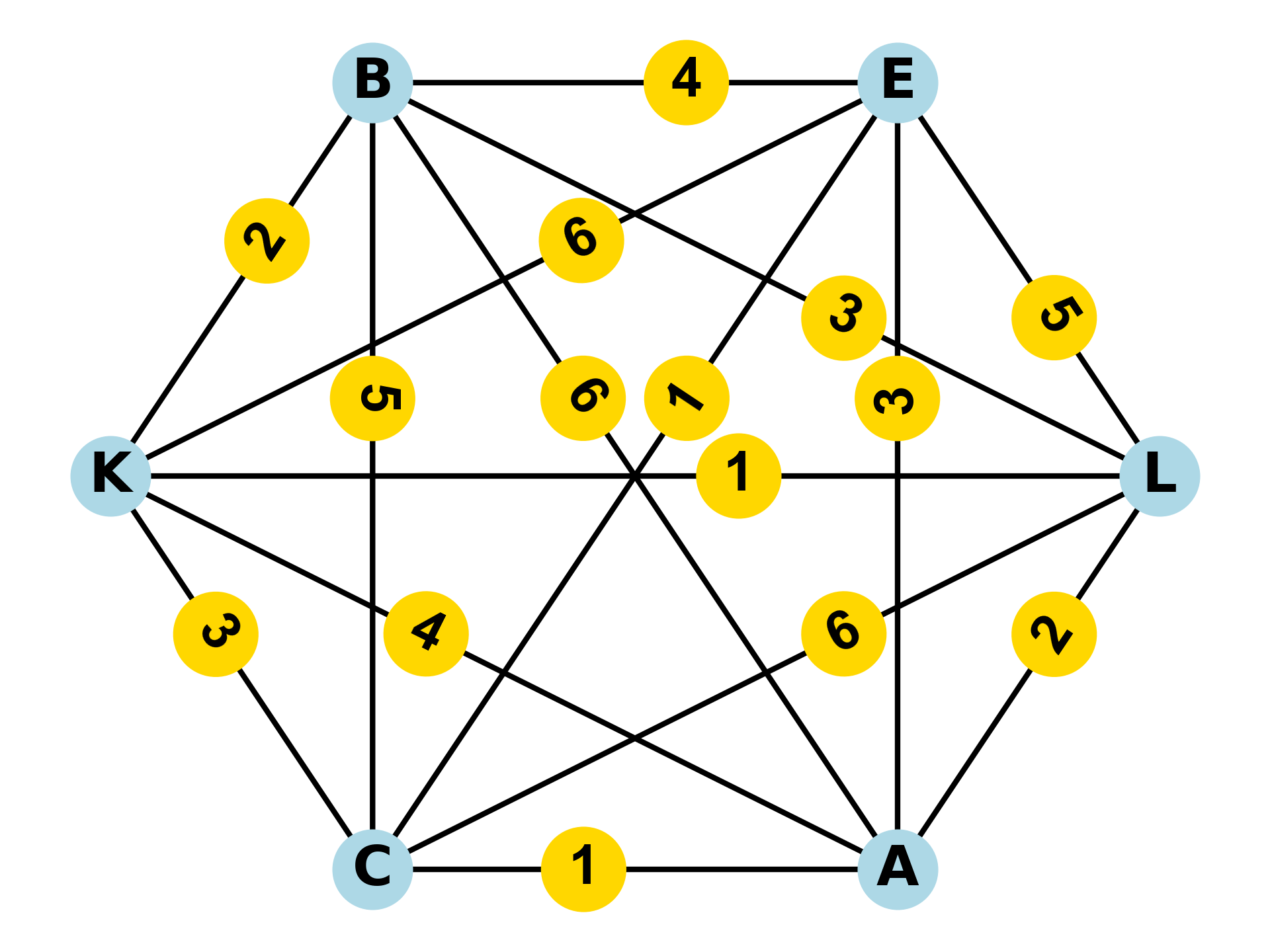} & \includegraphics[scale=\boardvisscale,valign=c]{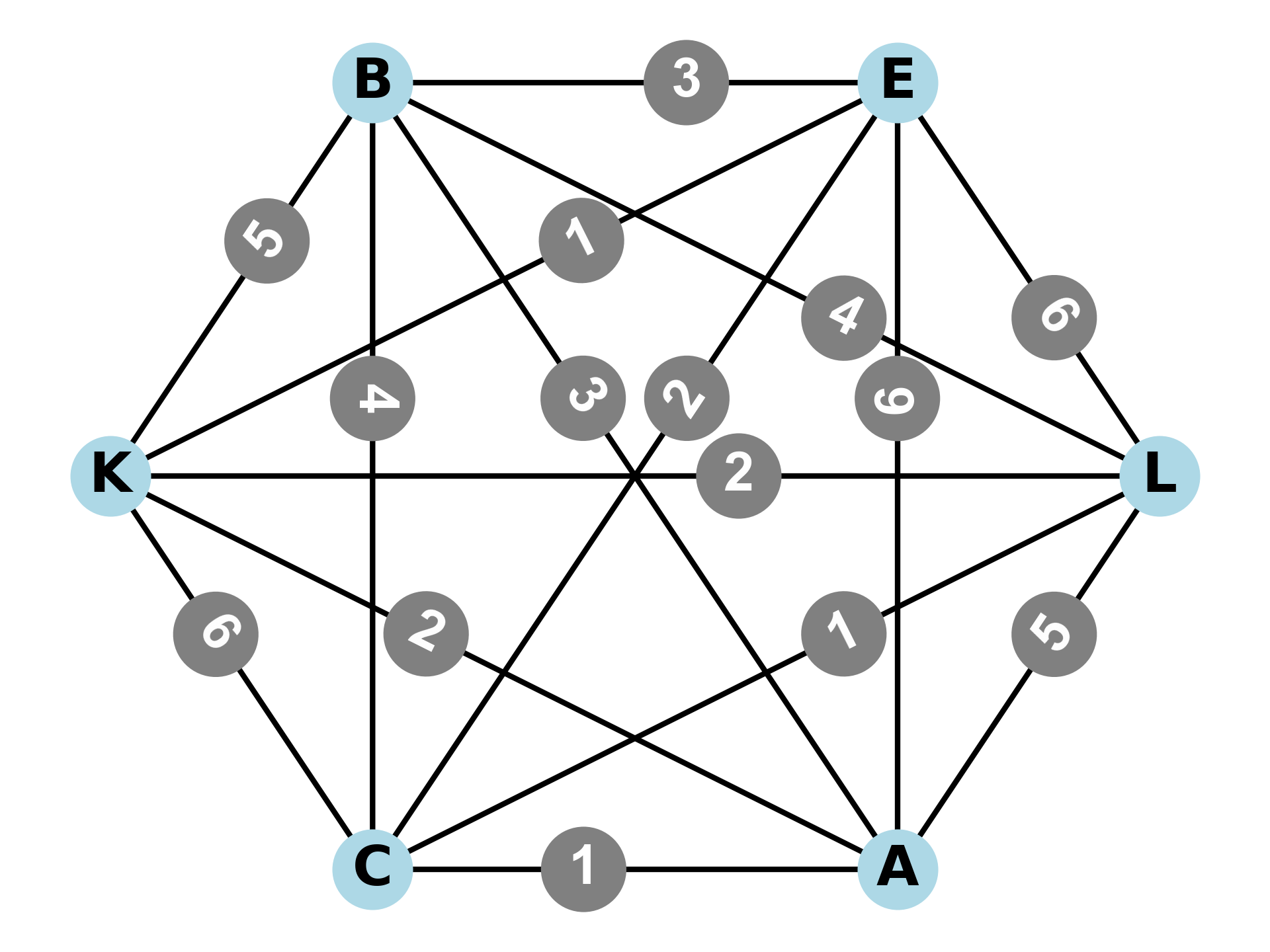} \\
        Board Setup 2 & \includegraphics[scale=\boardvisscale,valign=c]{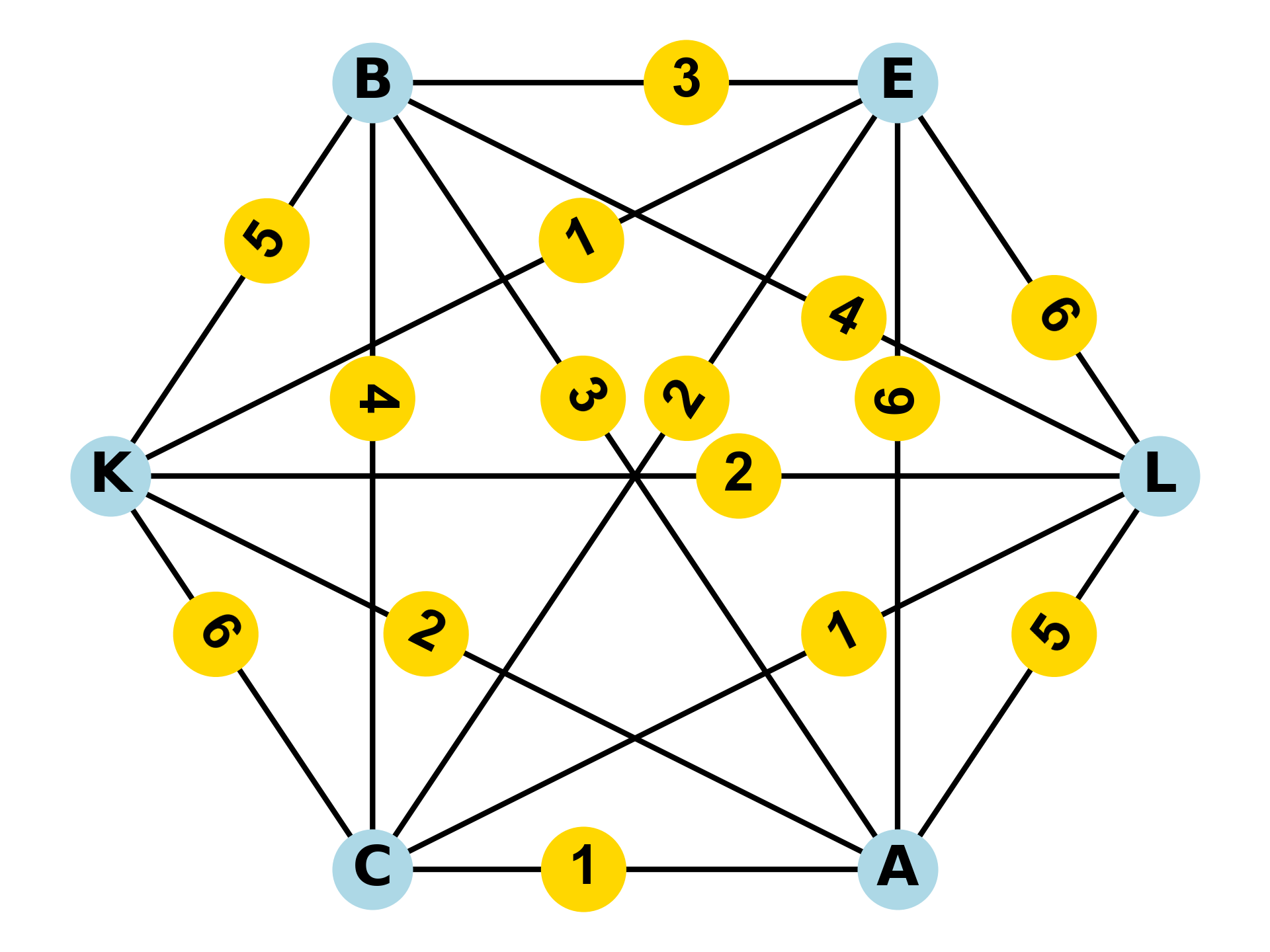} & \includegraphics[scale=\boardvisscale,valign=c]{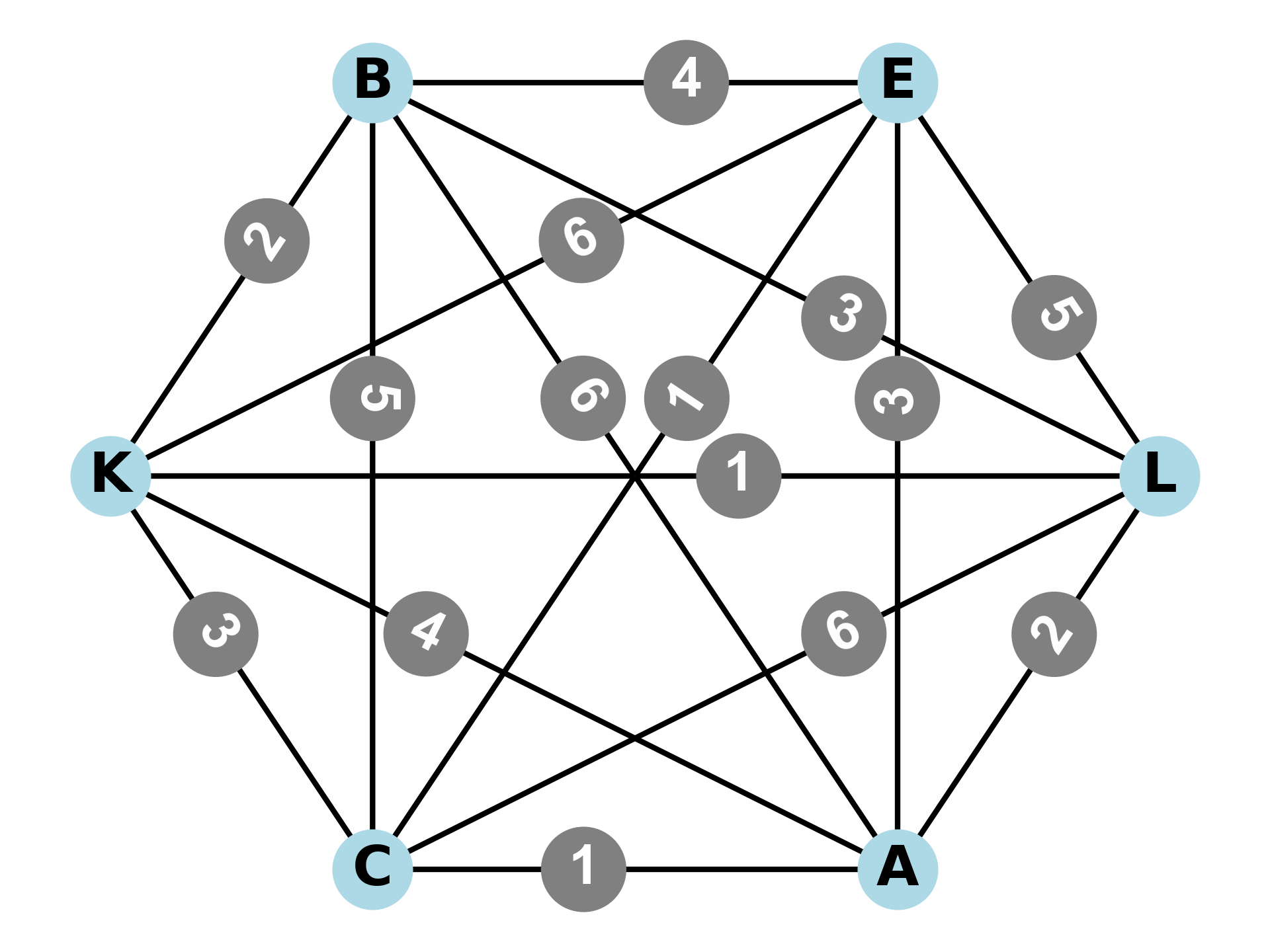} \\
        Board Setup 3 & \includegraphics[scale=\boardvisscale,valign=c]{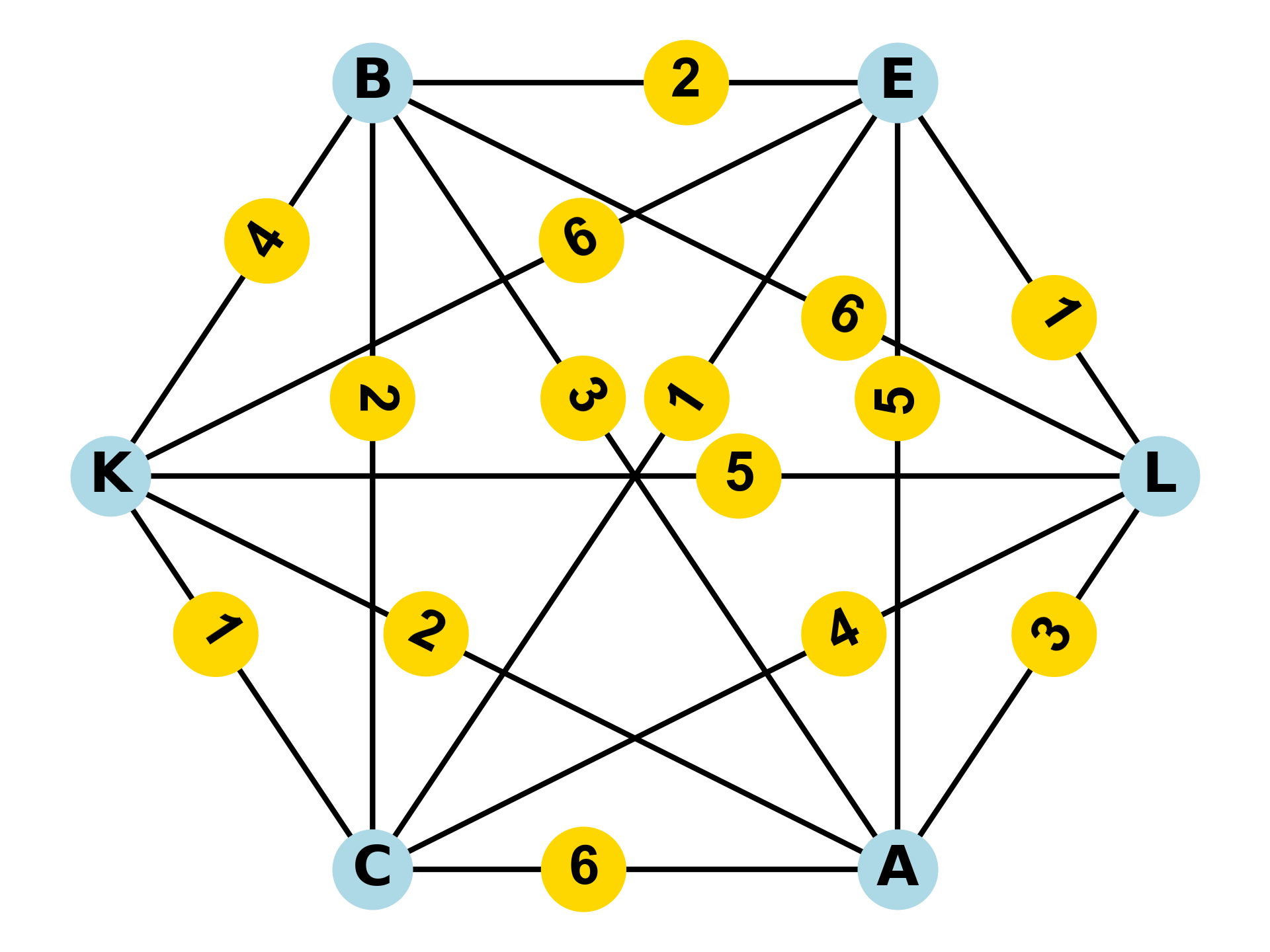} & \includegraphics[scale=\boardvisscale,valign=c]{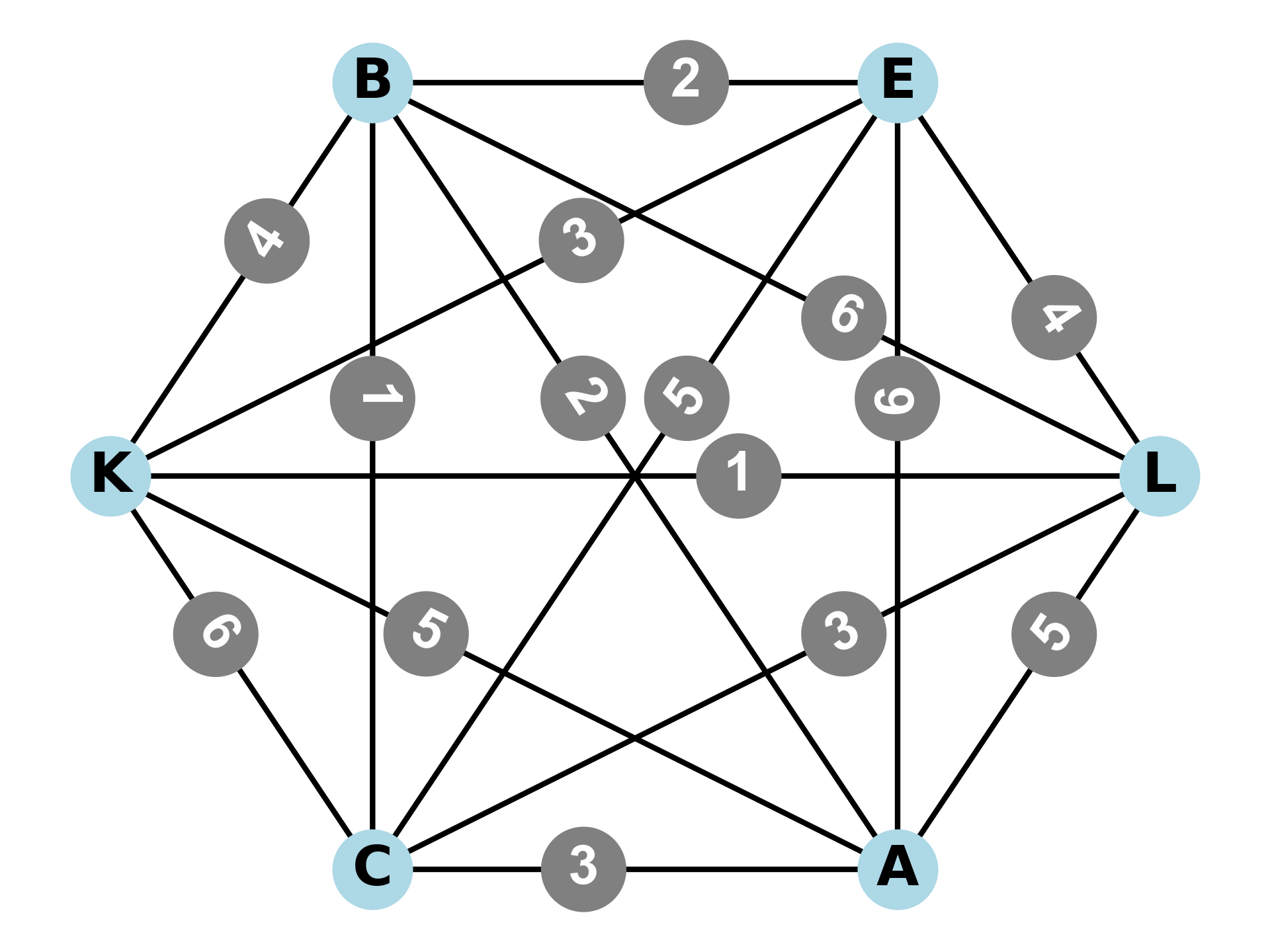} \\
        Board Setup 4 & \includegraphics[scale=\boardvisscale,valign=c]{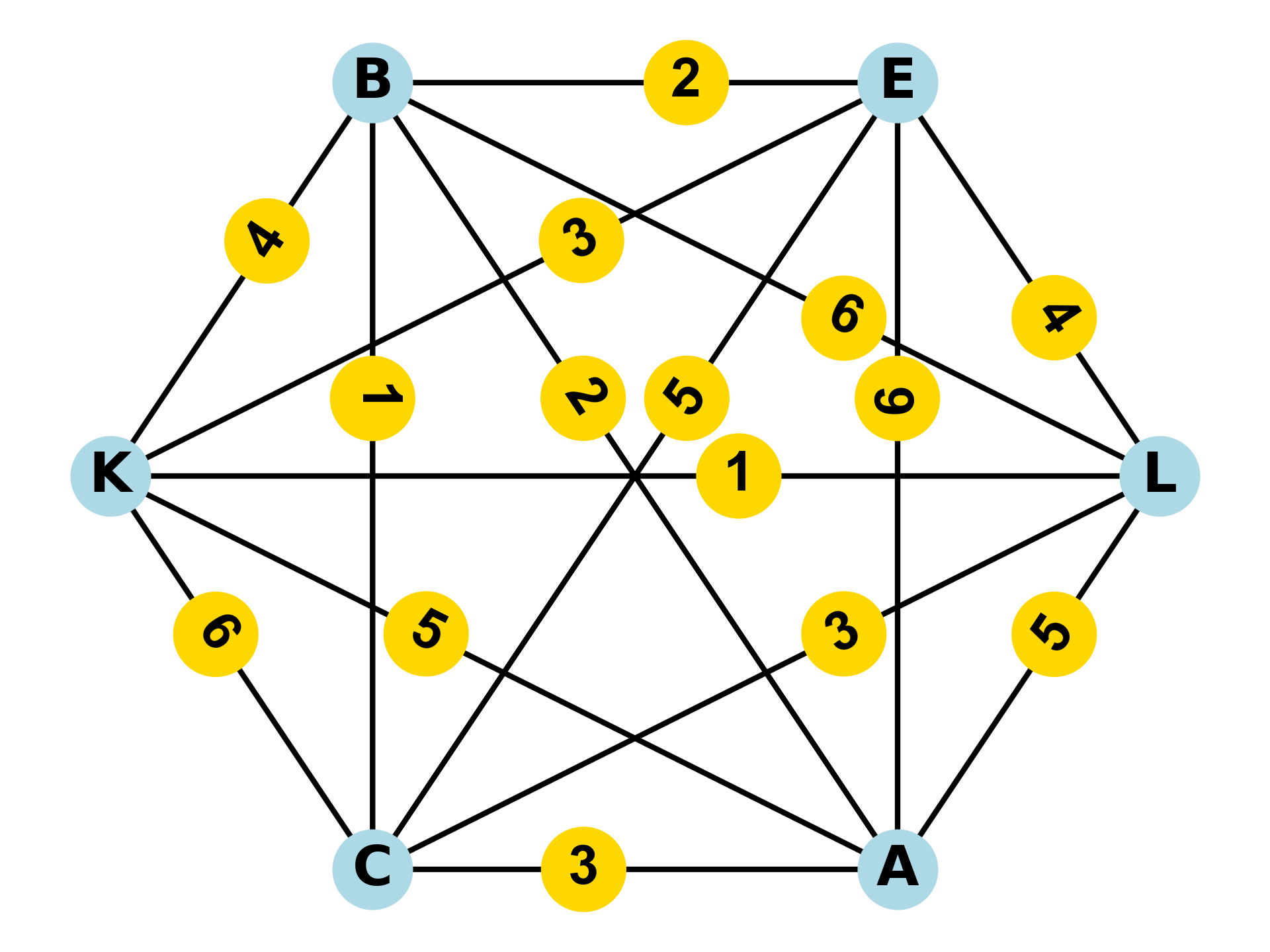} & \includegraphics[scale=\boardvisscale,valign=c]{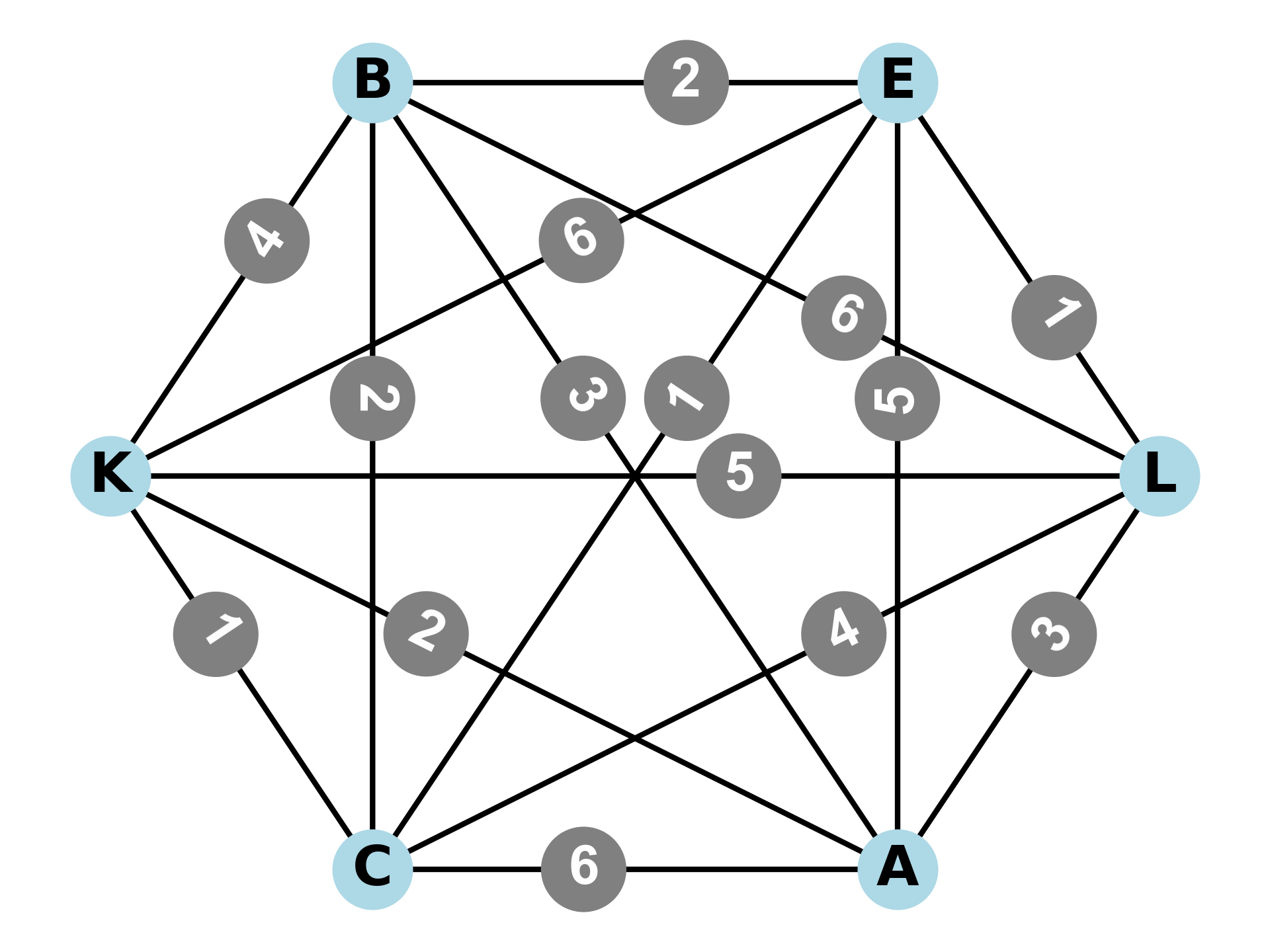} \\
        Board Setup 5 & \includegraphics[scale=\boardvisscale,valign=c]{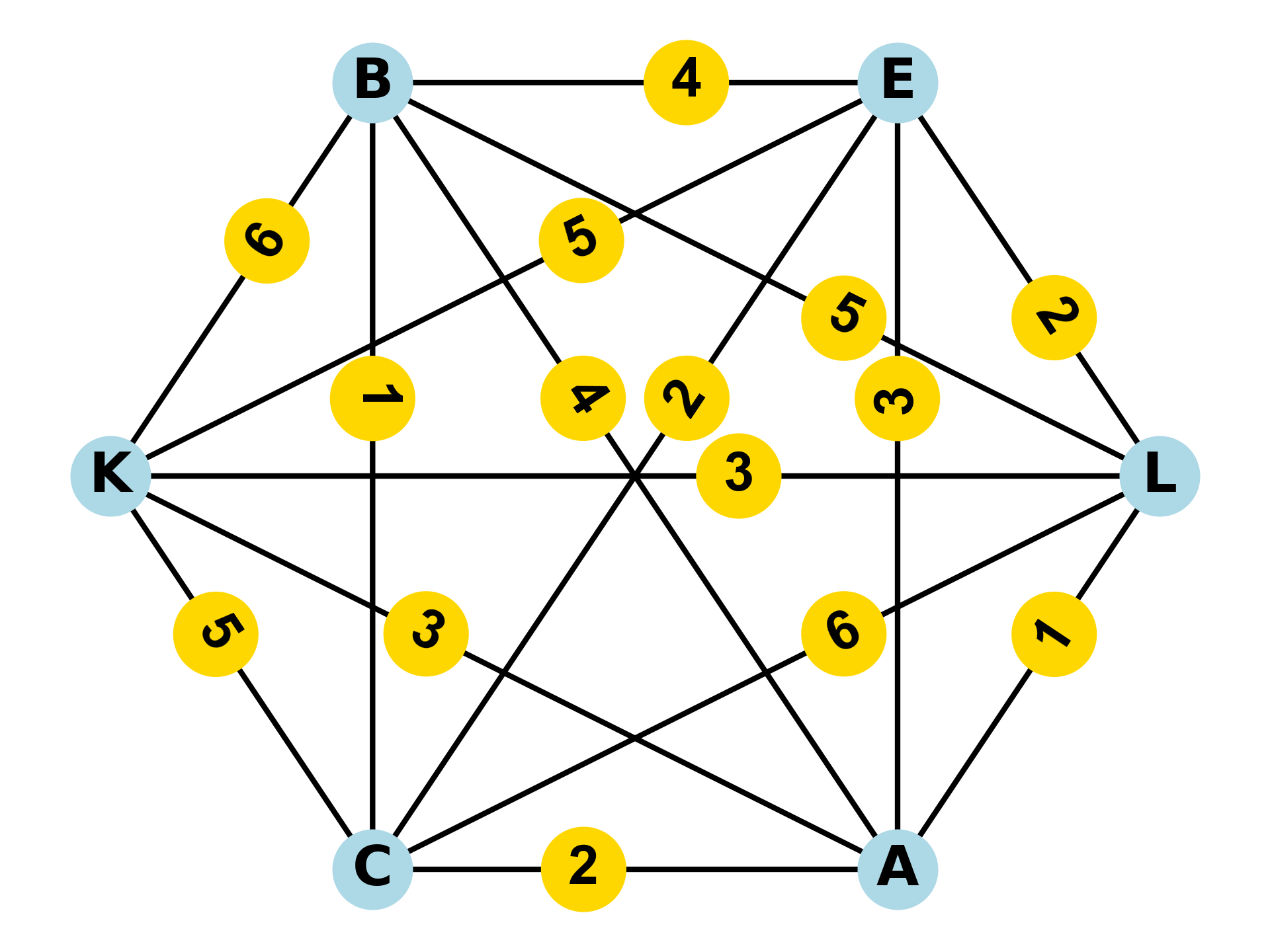} & \includegraphics[scale=\boardvisscale,valign=c]{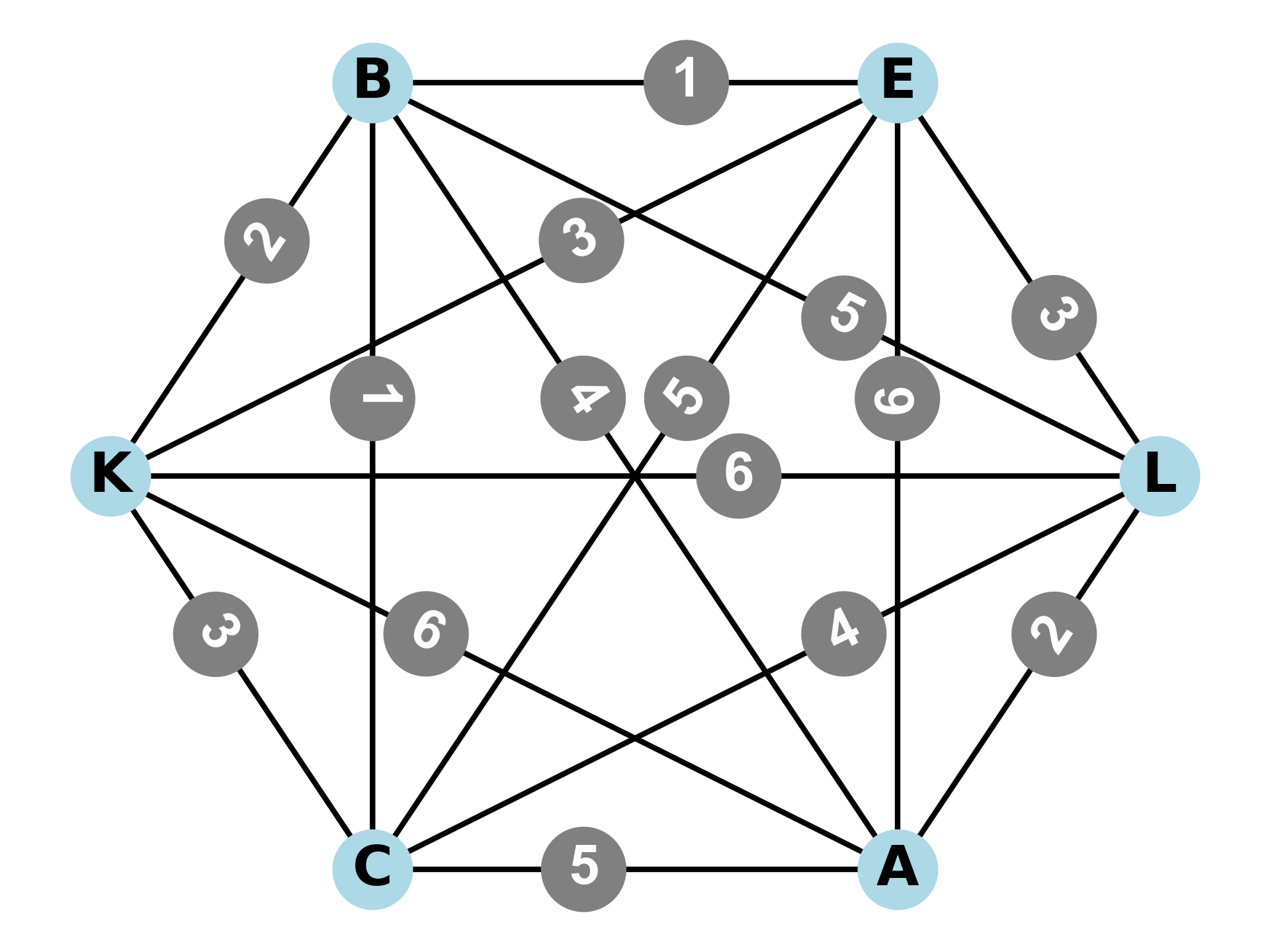} \\
        Board Setup 6 & \includegraphics[scale=\boardvisscale,valign=c]{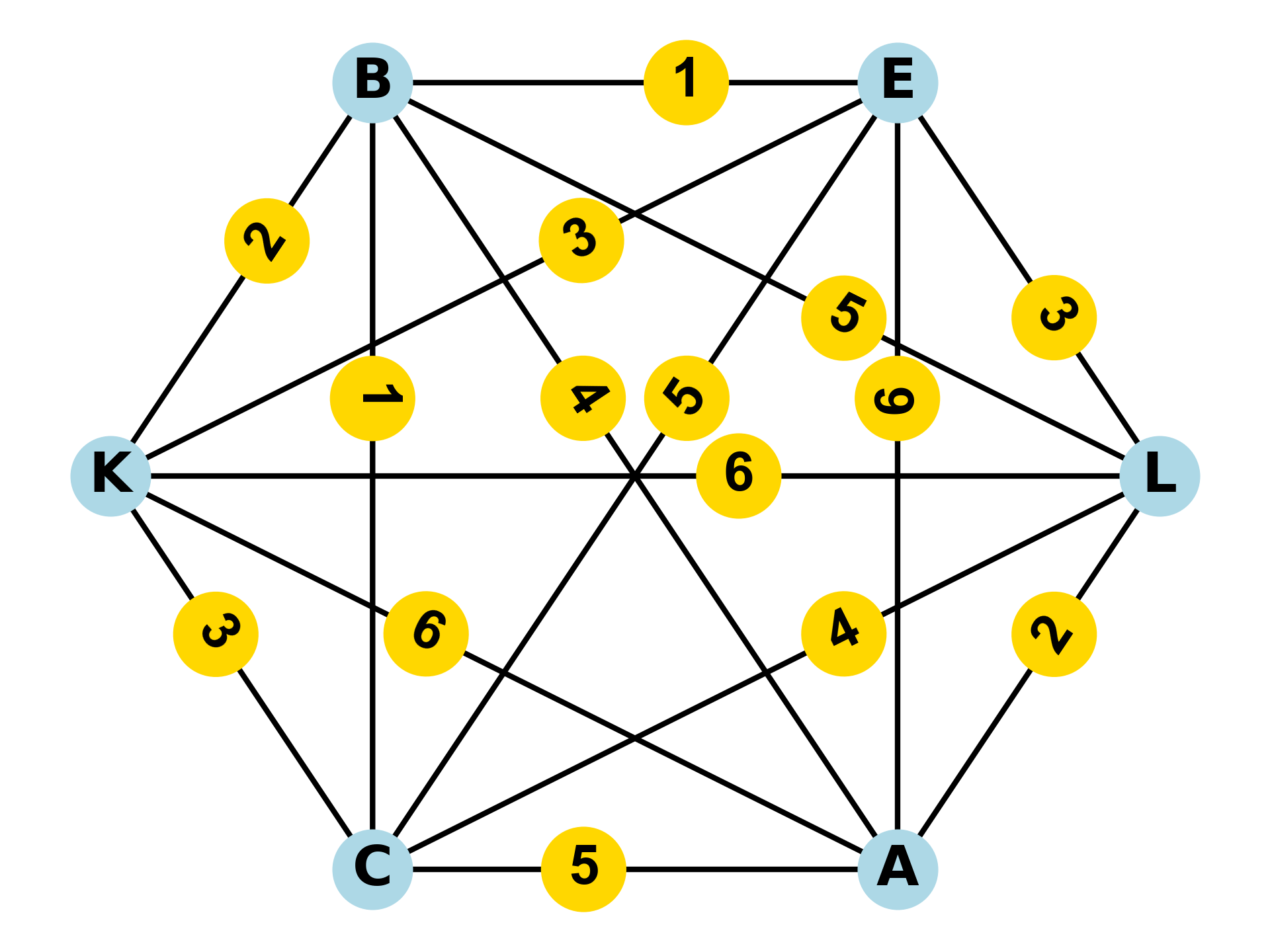} & \includegraphics[scale=\boardvisscale,valign=c]{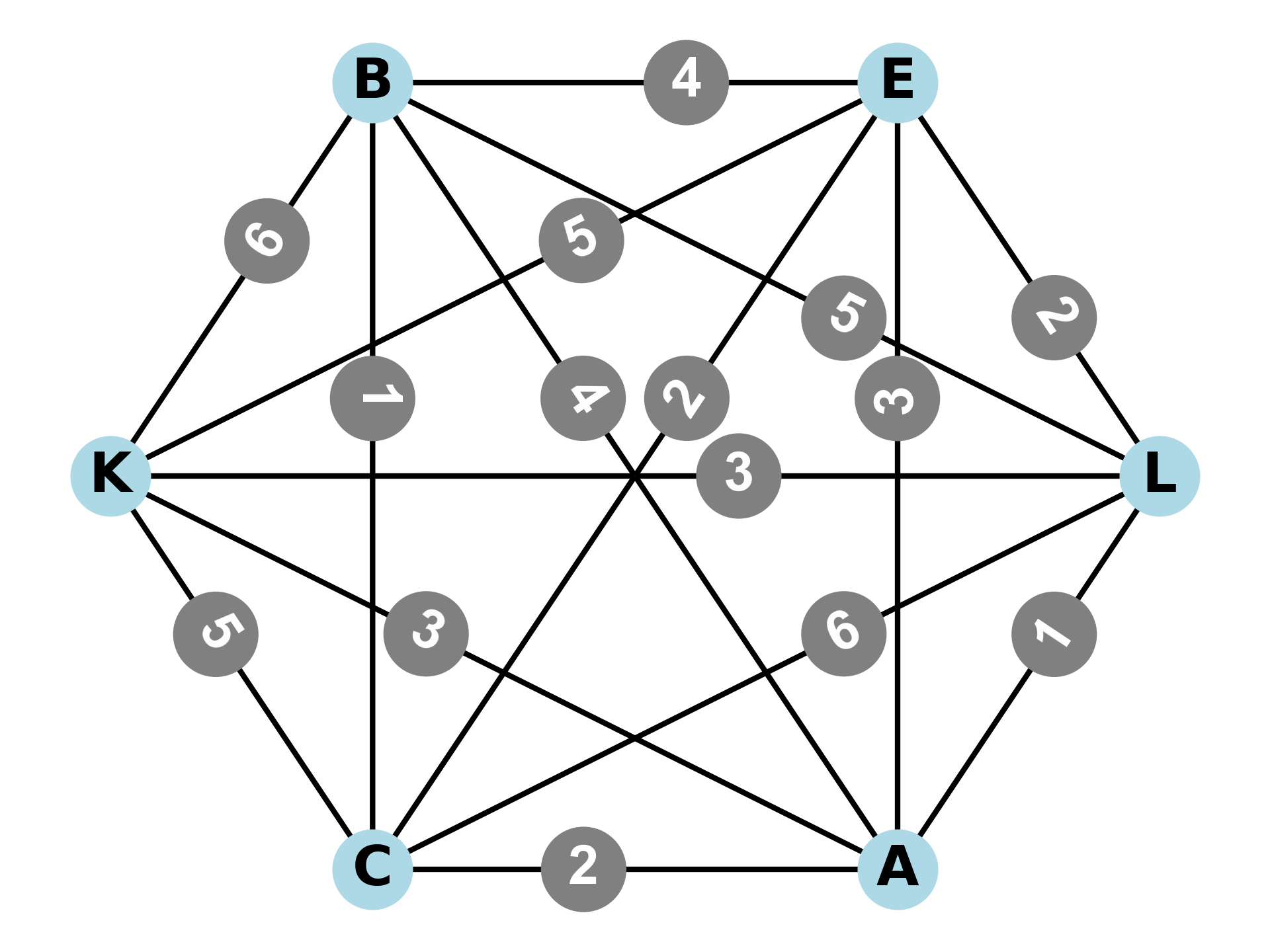} \\
    \end{tblr}
    \caption{The six board setups used in the experiments.}
    \label{tab:board-setups}
\end{table*}

\section{Full co-occurrence matrix}
\label{app:matrix}

\begin{figure*}[tbp]
    \centering
    \includegraphics[width=1\linewidth]{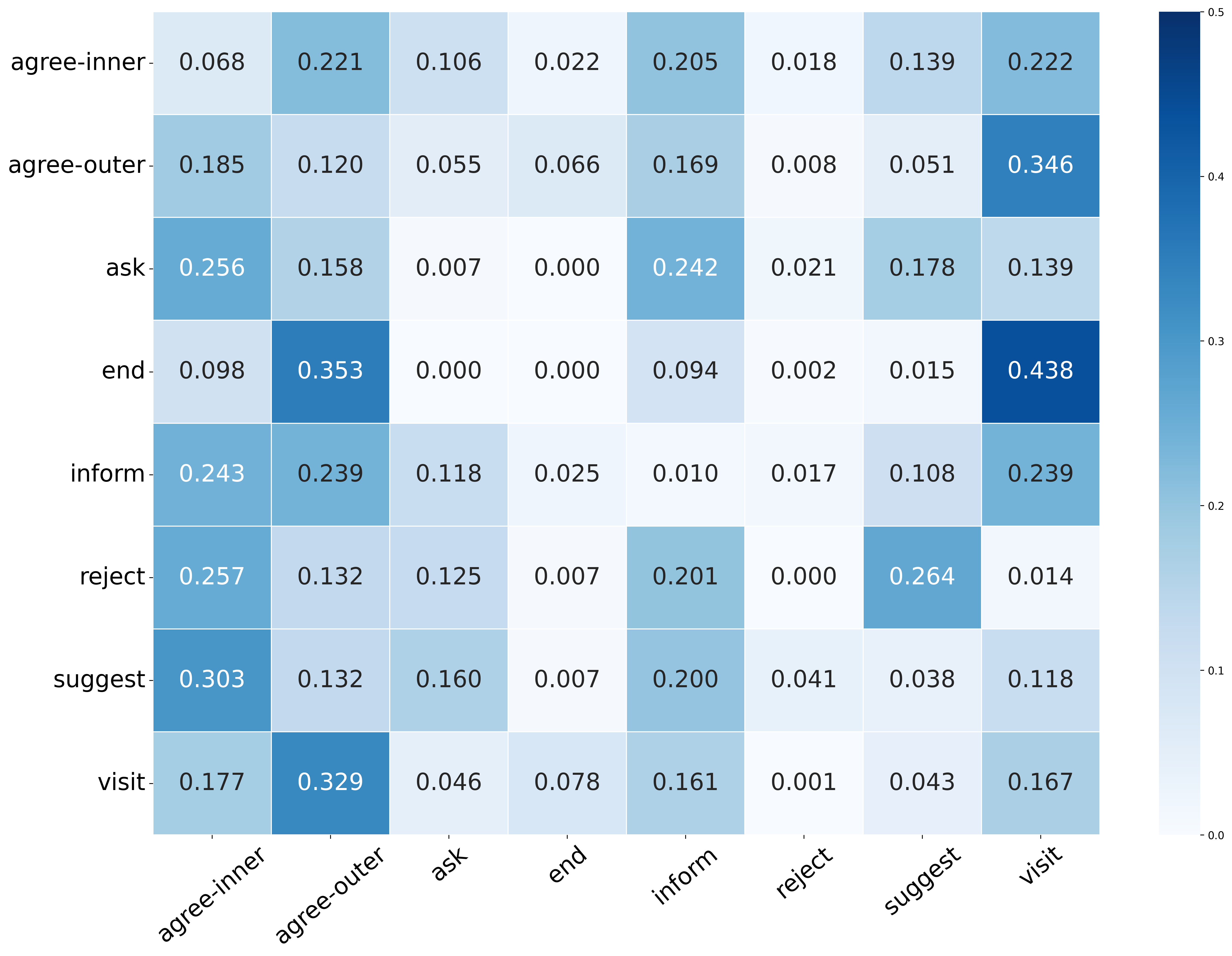}
    \caption{The full confusion matrix illustration the probabilities of each action along the y axis occurring with each action along the x axis; Problem-Solving \USER agent.}
    \label{fig:co-oc-full}
\end{figure*}

Figure \ref{fig:co-oc-full} illustrates the complete confusion matrix depicting the probabilities of each generated action along the y-axis occurring with each action along the x-axis. The figures represents results for the Problem-Solving \USER agent.

